\documentclass[journal,twoside,web]{ieeecolor}
\usepackage{generic}
\usepackage{cite}
\usepackage{url}
\usepackage{amsmath,amssymb,amsfonts}
\usepackage{graphicx}
\usepackage{textcomp}
 \usepackage{epstopdf}
 \usepackage{algorithm}
\usepackage{algpseudocode}
\usepackage{booktabs}
\usepackage{cases}
 \usepackage{array}
\usepackage{caption}
\usepackage{subfigure}
\usepackage{verbatim}

\def\BibTeX{{\rm B\kern-.05em{\sc i\kern-.025em b}\kern-.08em
    T\kern-.1667em\lower.7ex\hbox{E}\kern-.125emX}}
\def\QEDopen{{\setlength{\fboxsep}{0pt}\setlength{\fboxrule}{0.2pt}\fbox{\rule[0pt]{0pt}{1.3ex}\rule[0pt]{1.3ex}{0pt}}}} 
\def\QED{\QEDopen} 

\begin{document}
\title{An Entropy Weighted Nonnegative Matrix Factorization Algorithm for Feature Representation}
\author{Jiao Wei, Can Tong, Bingxue Wu, Qiang He, Shouliang Qi, Yudong Yao, \IEEEmembership{Fellow, IEEE}, and Yueyang Teng\textit{*}
\thanks{This work was supported by the Fundamental Research Funds for the Central Universities of China (N2019006 and N180719020).\\
\indent J. Wei, C. Tong, B. Wu, Q He and S. Qi are with the College of Medicine and Biological Information Engineering, Northeastern University, Shenyang 110169, China.\\
\indent Y. Yao is with the Department of Electrical and Computer Engineering, Stevens Institute of Technology, Hoboken, NJ 07030, USA.\\
\indent Y. Teng is with the College of Medicine and Biological Information Engineering, Northeastern University, Shenyang 110169, China, and also with the Key Laboratory of Intelligent Computing in Medical Image, Ministry of Education, Shenyang 110169, China (email: tengyy@bmie.neu.edu.cn).}}

\maketitle

\begin{abstract}
Nonnegative matrix factorization (NMF) has been widely used to learn low-dimensional representations of data. However, NMF pays the same attention to all attributes of a data point, which inevitably leads to inaccurate representation. For example, in a human-face data set, if an image contains a hat on the head, the hat should be removed or the importance of its corresponding attributes should be decreased during matrix factorizing. This paper proposes a new type of NMF called entropy weighted NMF (EWNMF), which uses an optimizable weight for each attribute of each data point to emphasize their importance. This process is achieved by adding an entropy regularizer to the cost function and then using the Lagrange multiplier method to solve the problem. Experimental results with several data sets demonstrate the feasibility and effectiveness of the proposed method.
We make our code available at \url{https://github.com/Poisson-EM/Entropy-weighted-NMF}.

\end{abstract}
\section{Introduction}
\IEEEPARstart{W}{ith} the rapid development of data acquisition technology, large amounts of data, such as online documents, medical images, various video series, traffic data, health data and other high-dimensional data, are accumulating. Therefore, dimensionality reduction \cite{ref1} has become an essential step in data mining. Vector quantization (VQ) \cite{ref2}, singular value decomposition (SVD) \cite{ref3}, principal component analysis (PCA) \cite{ref4}, independent component analysis (ICA) \cite{ref5} and concept factorization (CF) \cite{ref6}, \cite{ref7} are some most commonly used dimensionality reduction methods. However, due to negative components, these methods typically cannot be reasonably explained in some practical problems. Therefore, developing a nonnegative factorization method is valuable for research. Thus, researchers have investigated the nonnegative matrix factorization (NMF) proposed by Lee and Seung \cite{ref8}. This method represents a nonnegative matrix by a product of two low rank nonnegative matrices and then learns a parts-based representation. In recent years, NMF has been applied in many fields, including cluesting, dimensionality reduction, blind source separation, etc. \cite{ref9,ref10,ref11,ref12}.\\
\indent In recent years, many variants of NMF have been proposed to extend its applicable range. For example, Ding \emph{et al.} \cite{ref13} learned new low-dimensional features from data with convenient clustering interpretation using Semi-NMF, which allows the data matrix and the base matrix to have mixed signs \cite{ref14,ref15}. They also developed Convex-NMF by restricting the base vectors to a convex combination of the data points. Considering that the orthogonality constraint leads to sparsity, Pompili \emph{et al.} \cite{ref16} proposed an orthogonal NMF (ONMF) method, which adds orthogonality constraints to the base and representation matrices. There are two ways to solve ONMF: the Lagrange method \cite{ref16} and the natural gradient on the Stiefel manifold \cite{ref17}. \\

Blondel \emph{et al.} \cite{ref18} conducted an extended study for NMF by incorporating predetermined weights to each attribute of each data point, demonstrating that weights can produce important flexibility by better emphasizing certain features in image approximation problems. Other researchers then conducted a series of studies with the weighted NMF (WNMF). For example, Kim and Choi \cite{ref19} proposed a new WNMF method to process an incomplete data matrix with missing entries, which combined the binary weights into the NMF multiplication update. Li and Wu \cite{ref20} proposed a weighted nonnegative matrix tri-factorization method for co-clustering that weights each row and column of the original matrix in a specific way, and the normalized cut information is combined into the optimization model. Then, to assign labels to images, Kalayeh \emph{et al.} \cite{ref21} proposed a weighted expansion method of multiview NMF, which imposes a consistent constraint on the representation matrices between different features, and a weight matrix mitigates data set imbalance. Dai \emph{et al.} \cite{ref22} proposed applying the WNMF method to image recovery, and experimental results showed that, particularly for data affected by salt and pepper noise, the method could remove noise effectively and could also provide a more accurate subspace representation. \\

 We refer these methods as ``hard WNMF''; however, one primary drawback is that the weights, which they rely on, must be predetermined. Variable weights for NMF are challenging to implement that can be resolved by designing interpretable and computable weights.

This paper presents an entropy weighted NMF (EWNMF) method that assigns a weight to indicate the importance of each attribute of each data point in matrix factorizing. Then, the entropy of these weights is used to regularize the cost function for obtaining an easy computable solution, which makes the range of weights fall within [0, 1] with a summation of one; thus, the weights can be explained as the probability of the contribution of an attribute of one data point to NMF. Experimental results with several real data sets show that EWNMF performs better than other NMF variants.

\section{Methodology}
\subsection{Related research}
\indent NMF is a matrix factorization method that focuses on data matrices with nonnegative elements, which can reveal hidden structures and patterns from generally redundant data. We will review the standard NMF as follows.\\
$\mathbf{Notations}~$~In this paper, matrices are denoted as capital letters. For a matrix $A$, $A_{*i}$, $A_{i*}$ and $A_{ij}$ denote the $i-th$ column, the $i-th$ row and $(i,j)-th$ element of $A$, respectively; the Frobenius norm is represented as $ \left\|A\right\|_F $; $\odot$ and $. /$ mean the item-by-item multiplication and division of two matrices, respectively; $A^T$ denotes the transpose of $A$; $A \geq 0$ means that all elements of $A$ are equal to or larger than 0.\\
\indent The expression of NMF is:
\begin{equation}
	X\approx W H
	\label{eq1}
\end{equation}
where the matrix $X\subseteq R^{M \times N}$ denotes the given nonnegative matrix in which each column is a data point. The goal of NMF is to find two low-dimensional nonnegative matrices: $W\subseteq R^{M \times K}$ is called the base matrix, and $H\subseteq R^{K \times N}$ is called the representation matrix, whose product can approximate the original matrix \cite{ref23,ref24}, where $K<<min\{M,N\}$. \\
\indent There are different standards to measure the quality of decomposition. Lee and Seung proposed using the square of the Euclidean distance and the Kullback-Leibler divergence. In this paper, the Euclidean distance is used, and the formula is expressed as:
\begin{equation}
\begin{aligned}
    \min&~F_1(W,H)=\left\|X-WH\right\|_F^2 \\
    s.~t.&~W\geq 0,~H \geq 0
    \label{eq2}
\end{aligned}
\end{equation}
\indent To alternatively minimize $W$ and $H$ in Eq. (\ref{eq2}), the construction of the auxiliary function is important to determine the iterative update rule.\\
$\mathbf{Definition~1} ~(Auxiliary~function)$ If the function $G(h,h')$ satisfies the following conditions:
\begin{equation}
	\begin{aligned}
		&G(h,h')\ge F(h)\\
		&G(h',h')=F(h')
		\label{eq3}
	\end{aligned}    
\end{equation}
where $h'$ is a given value; then, $G(h,h')$ is the auxiliary function of $F(h)$ on $h'$. 

Then, we can draw the following conclusion.\\
$\mathbf{Lemma~1.}~$ If $G(h,h')$ is an auxiliary function of $F(h)$, then under the update rule:
\begin{equation}
	h^*=arg\mathop{\min}_{h}G(h,h')
	\label{eq4}
\end{equation}
the function $F(h)$ does not increase. \\
$\mathbf{Proof.~}$ The conditions satisfied by the auxiliary function make this proof marked because:
\begin{equation}
	F(h^{*})\leq G(h^{*},h')\leq G(h',h')\leq F(h')
	\label{eq5}
\end{equation}
\rightline{$${\boxed{}}$$}
\indent Thus, if the auxiliary function reaches the minimum, the original function should decrease.\\
\indent Then, we construct the update rule for the NMF problem. We consider $W$ first, where $W^t>0$ and $H>0$ are given. Let $\xi_{ijk}=W^t_{ik}H_{kj}/(W^tH)_{ij}$, of course, $\xi_{ijk}\geq0$ and $\sum_{k=1}^K\xi_{ijk}=1$. Therefore, the auxiliary function of standard NMF is:
\begin{equation}
	f_1(W,W^t)=\sum_{i=1}^M\sum_{j=1}^N\sum_{k=1}^K\xi_{ijk}(X_{ij}-\frac{W_{ik}H_{kj}}{\xi_{ijk}})^2
	\label{eq6}
\end{equation}
\indent Because the function is separable, it can be easily minimized. We thus take the partial derivative of Eq. (\ref{eq6}) and set it to zero so that we can obtain the following update rule as:
\begin{equation}
    W \leftarrow W \odot (XH^T) ~./~ (WHH^T)
    \label{eq7}
\end{equation}
\indent The method of constructing the auxiliary function of $H$ is similar to that of $W$. Then, the following update of $H$ is obtained:
\begin{equation}
    H \leftarrow H \odot (W^TX) ~./~ (W^TWH)
    \label{eq8}
\end{equation}
\begin{figure}[!t]  
\centering
\includegraphics[width=\columnwidth]{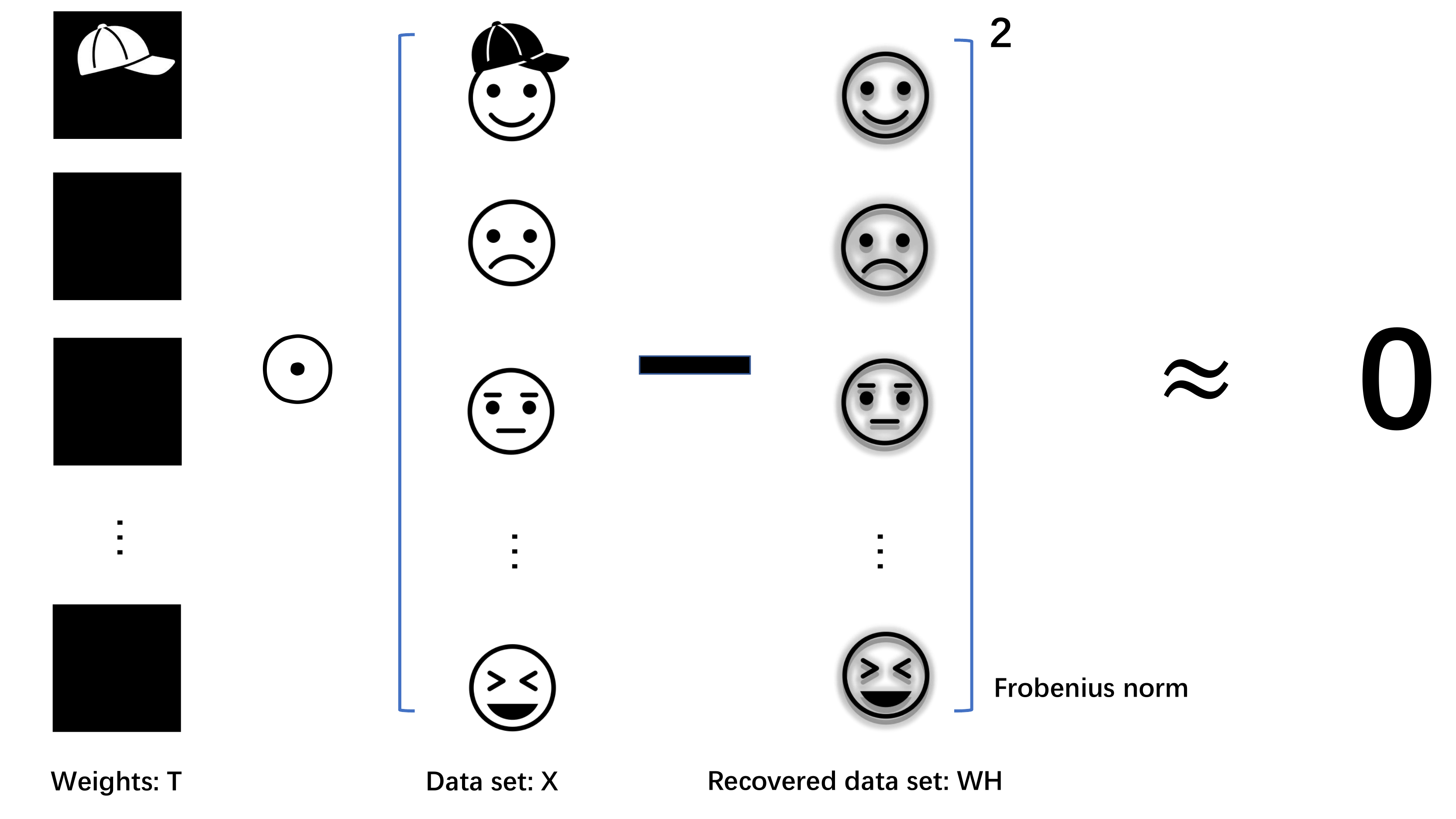}
\caption{Diagram of the proposed method: a hat in the first image destroys the accuracy of feature representation, which should be assigned no or little importance.}
\label{fig:1}
\end{figure}
\subsection{Proposed method}
\indent Different from previous methods, an optimizable weight matrix is used to measure the importance of the attributes in matrix factorizing. Fig. \ref{fig:1} shows the explanation of the proposed idea that, if there is a hat in the only image, the corresponding attributes of the hat must destroy the result of NMF, and we should certainly eliminate or weaken their importances. Thus, it may be necessary to provide different importances to the attributes of each data point based on the following constraint, which will lead to a new optimization model:
\begin{equation}
\begin{aligned}
 \min&~F_2(W,H,T)=\sum_{i=1}^M \sum_{j=1}^NT_{ij}[X_{ij}-(WH)_{ij}]^2\\ 
 s.~t.&~ W\geq0,~H\geq0,~T\geq0,~\sum_{i=1}^MT_{ij}=1
    \label{eq9}
\end{aligned}
\end{equation}
\indent We first consider the new variable $T$, which can be solved in an alternative optimization manner. For fixed $W$ and $H$, $T_{ij}$ is very easy to solve as $T_{ij}=1$ if $E_{ij}=min\{|E_{1j}|, |E_{2j}|,..., |E_{Mj}|\}$ or 0 otherwise$\footnote{The update rule can be easily explained by an example as: \begin{equation}
\begin{aligned}
 \min~\{3, 1, 2\}= min&~3T_1+1T_2+2T_3\\ 
 s.~t.&~ T_1\geq0, T_2\geq0, T_3\geq0\\
 &~T_1+T_2+T_3=1\nonumber
\end{aligned}
\end{equation} The solution is that $T_1=0$, $T_2=1$ and $T_3=0$, in which $T_2$ corresponds to the minimum value of \{3, 1, 2\}. This process is similar to the computation of the weights in the k-means algorithm.}$, where $E=X-WH$. And it demonstrates the simple fact that only one element of $T_{*j}$ is 1, and the others are 0, which is incompatible with the real problem.\\
\indent To address this issue, we apply an entropy regularizer to penalize the cost function of NMF to obtain a weight in the range of [0, 1] instead of 0 or 1, which uses information entropy to calculate the uncertainty of weights. The new optimization problem is rewritten as follows:\\
\begin{equation}
\begin{aligned}
 \min&~F_3(W,H,T)=\sum_{i=1}^M \sum_{j=1}^NT_{ij}[X_{ij}-(WH)_{ij}]^2\\
 &~~~~~~~~~~~~~~~~~~~+\gamma\sum_{i=1}^M\sum_{j=1}^NT_{ij}ln(T_{ij})\\ 
 s.~t.&~ W\geq0,~H\geq0,~T\ge0,~\sum_{i=1}^MT_{ij}=1
    \label{eq10}
\end{aligned}
\end{equation}
where $\gamma\ge0$ is a given hyperparameter. The first term in Eq. (\ref{eq10}) is the sum of errors, and the second term is the negative entropy of the weights. The original cost function in Eq. (\ref{eq9}) results in only one attribute of each data point being involved in feature representation, and the entropy regularizer will stimulate more attributes to help feature representation. \\
\indent This equation can be solved by a simple algorithm, which is based on the following proposition:\\
$\mathbf{Proposition~1.~}$Given the matrices $W$ and $H$, $T_{ij}$ in Eq. (\ref{eq10}) is minimized when:
\begin{equation}
  T_{ij}=\frac{e^{-\frac{[X_{ij}-(WH)_{ij}]^2}{\gamma}}}{\sum_{l=1}^Me^{-\frac{[X_{lj}-(WH)_{lj}]^2}{\gamma}}}
    \label{eq11}
\end{equation}
$\mathbf{Proof.~}$
We construct the Lagrange function of Eq. (\ref{eq10}) with respective to $T$ as:\\
\begin{equation}
\begin{aligned}
L(T,\lambda)=&\sum_{i=1}^M \sum_{j=1}^NT_{ij}[X_{ij}-(WH)_{ij}]^2\\
 &+\gamma\sum_{i=1}^M\sum_{j=1}^NT_{ij}ln(T_{ij})-\sum_{j=1}^N\lambda_j(\sum_{i=1}^MT_{ij}-1)
    \label{eq12}
 \end{aligned}
\end{equation}
where $[\lambda_1,\lambda_2,...,\lambda_N]$ is a vector containing the Lagrange multipliers corresponding to the constraints.\\
\indent By setting the gradient of Eq. (\ref{eq12}) with respect to $\lambda_j$ and $T_{ij}$ to zero, we obtain the following equation system:\\
\begin{numcases}{}
	\frac{\partial L}{\partial \lambda_j}&=$\sum_{i=1}^MT_{ij}-1=0$ \label{eqsystem1}\\
	\frac{\partial L}{\partial T_{ij}}&=$[X_{ij}-(WH)_{ij}]^2+\gamma lnT_{ij}+\gamma-\lambda_j=0$
\end{numcases}
From Eq. (14), we know that:\\
\begin{equation}
\begin{aligned}
T_{ij}=e^{\frac{\lambda_j-\gamma}{\gamma}}e^{-\frac{[X_{ij}-(WH)_{ij}]^2}{\gamma}}
    \label{eq15}
    \end{aligned}
\end{equation}
Substituting Eq. (\ref{eq15}) into Eq. (13), we have:\\
\begin{equation}
\begin{aligned}
\sum_{i=1}^MT_{ij}=e^{\frac{\lambda_j-\gamma}{\gamma}}\sum_{i=1}^M e^{{-\frac{[X_{ij}-(WH)_{ij}]^2}{\gamma}}}=1 
    \label{eq16}
    \end{aligned}
\end{equation}
It follows that:\\
\begin{eqnarray}
e^{\frac{\lambda_j-\gamma}{\gamma}}     =\frac{1}{\sum_{l=1}^M e^{{-\frac{[X_{lj}-(WH)_{lj}]^2}{\gamma}}}}  
    \label{eq17}
\end{eqnarray}
Substituting this expression to Eq. (\ref{eq15}), we find that:\\
\begin{equation}
T_{ij}=\frac{e^{-\frac{[X_{ij}-(WH)_{ij}]^2}{\gamma}}}{\sum_{l=1}^M e^{-\frac{[X_{lj}-(WH)_{lj}]^2}{\gamma}}} 
    \label{eq18}
\end{equation}
\rightline{$${\boxed{}}$$}
\indent Then, we can solve $W$ and $H$ with fixed $T$, which is similar to the standard NMF method. For example, we can construct the following auxiliary function about $W$:\\
\begin{equation}
\begin{aligned}
f_3(W,W^t)=&\sum_{i=1}^M \sum_{j=1}^N\sum_{k=1}^KT_{ij}\xi_{ijk}(X_{ij}-\frac{W_{ik}H_{kj}}{\xi_{ijk}})^2\\
&+\gamma\sum_{i=1}^M\sum_{j=1}^NT_{ij}lnT_{ij}
    \label{eq19}
    \end{aligned}
\end{equation}
Setting the partial derivative of $f_3(W,W^t)$ to zero yields the following update rule:\\
\begin{equation}
W \leftarrow W \odot (T \odot X)H^T ~./~\{[T \odot (WH)]H^T\} 
    \label{eq20}
\end{equation}
Similarly, we can also easily obtain the update rule for $H$ as follows:\\
\begin{equation}
H \leftarrow H \odot W^T(T \odot X)~./~\{W^T[T\odot(WH)]\} 
    \label{eq21}
\end{equation}
\indent The update rules to $W$ and $H$ are similar to the existing WNMF methods in \cite{ref21}, \cite{ref22}. Optimizing EWNMF is summarized as follows in $\mathbf{Algorithm~ 1}$:\\
 \begin{algorithm}[htb]
  \caption{ EWNMF}
  \label{alg:Framwork}
  \begin{algorithmic}[1]
    \Require
     Given the input nonnegative matrix $X\subseteq R^{M \times N}$, the number
\emph{K} of reduced dimensions and hyperparameter $\gamma$;
    \Ensure
     the weight matrix $T$, the base matrix $W$ and the representation matrix $H$;
    \State Randomly initialize $W\subseteq R^{M \times K}\textgreater 0$ and $H\subseteq R^{K \times N} \textgreater 0$;
    \While{not convergence}
    \label{code:fram:trainbase}
    \State Update $T$ by Eq. (\ref{eq18});
    \label{code:fram:add}
    \State Update $W$ by Eq. (\ref{eq20});
    \label{code:fram:classify}
    \State Update $H$ by Eq. (\ref{eq21});
    \label{code:fram:classify}
    \EndWhile
    \label{code:fram:select} \\
    \Return  $T$, $W$ and $H$.
  \end{algorithmic}
\end{algorithm}
\subsection{Extensions}
\indent The proposed entropy weighted method can also be applied to standard NMF using the square of the Euclidean distance and to those that use KL divergence, $\alpha$-divergence and other existing NMFs, such as ONMF, Semi-NMF, Convex-NMF, etc., which demonstrates its good compatibility. \\
\indent For example, incorporating the entropy regularizer into KL-NMF, its cost function is expressed as:
\begin{equation}
	\begin{aligned}
		F_4(T,W,H)=&\sum_{i=1}^M \sum_{j=1}^NT_{ij}[X_{ij}log\frac{X_{ij}}{(WH)_{ij}}-X_{ij}\\
		&+(WH)_{ij}]+\gamma\sum_{i=1}^M\sum_{j=1}^NT_{ij}ln(T_{ij})
		\label{eq22}
	\end{aligned}
\end{equation}
where $T$ can be derived from the above method, and then the update is shown below:
\begin{equation}
	T_{ij}=\frac{e^{-\frac{X_{ij}log\frac{X_{ij}}{(WH)_{ij}}-X_{ij}+(WH)_{ij}}{\gamma}}}{\sum_{l=1}^Me^{-\frac{X_{lj}log\frac{X_{lj}}{(WH)_{lj}}-X_{lj}+(WH)_{lj}}{\gamma}}} 
	\label{eq23}
\end{equation}
\indent Applying the entropy regularizer to the $\alpha$-divergence-based NMF method, the cost function is:
\begin{equation}
	\begin{aligned}
		F_5(T,W,H)=&\frac{1}{\alpha(\alpha-1)}\sum_{i=1}^M \sum_{j=1}^NT_{ij}[X_{ij}^{\alpha}(WH)_{ij}^{1-\alpha}-\alpha X_{ij}\\
		&+(\alpha-1)(WH)_{ij}]+\gamma\sum_{i=1}^M\sum_{j=1}^NT_{ij}ln(T_{ij}) 
		\label{eq24}
	\end{aligned}
\end{equation}
where $\alpha \in R$ is a given value. The update method of $T$ is as follows:
\begin{equation}
	T_{ij}=\frac{e^{-\frac{[X_{ij}^{\alpha}(WH)_{ij}^{1-\alpha}-\alpha X_{ij}+(\alpha-1)(WH)_{ij}]}{\alpha(\alpha-1)\gamma}}}{\sum_{l=1}^Me^{-\frac{[X_{lj}^{\alpha}(WH)_{lj}^{1-\alpha}-\alpha X_{lj}+(\alpha-1)(WH)_{lj}]}{\alpha(\alpha-1)\gamma}}} 
	\label{eq25}
\end{equation}
\indent The entropy weighted method can also be popularized to many other NMFs, which is similar to the above derivations, and thus, we omit them.
\begin{figure*}[htbp]
	\centering
	\subfigure[]{
		\includegraphics[scale=0.4]{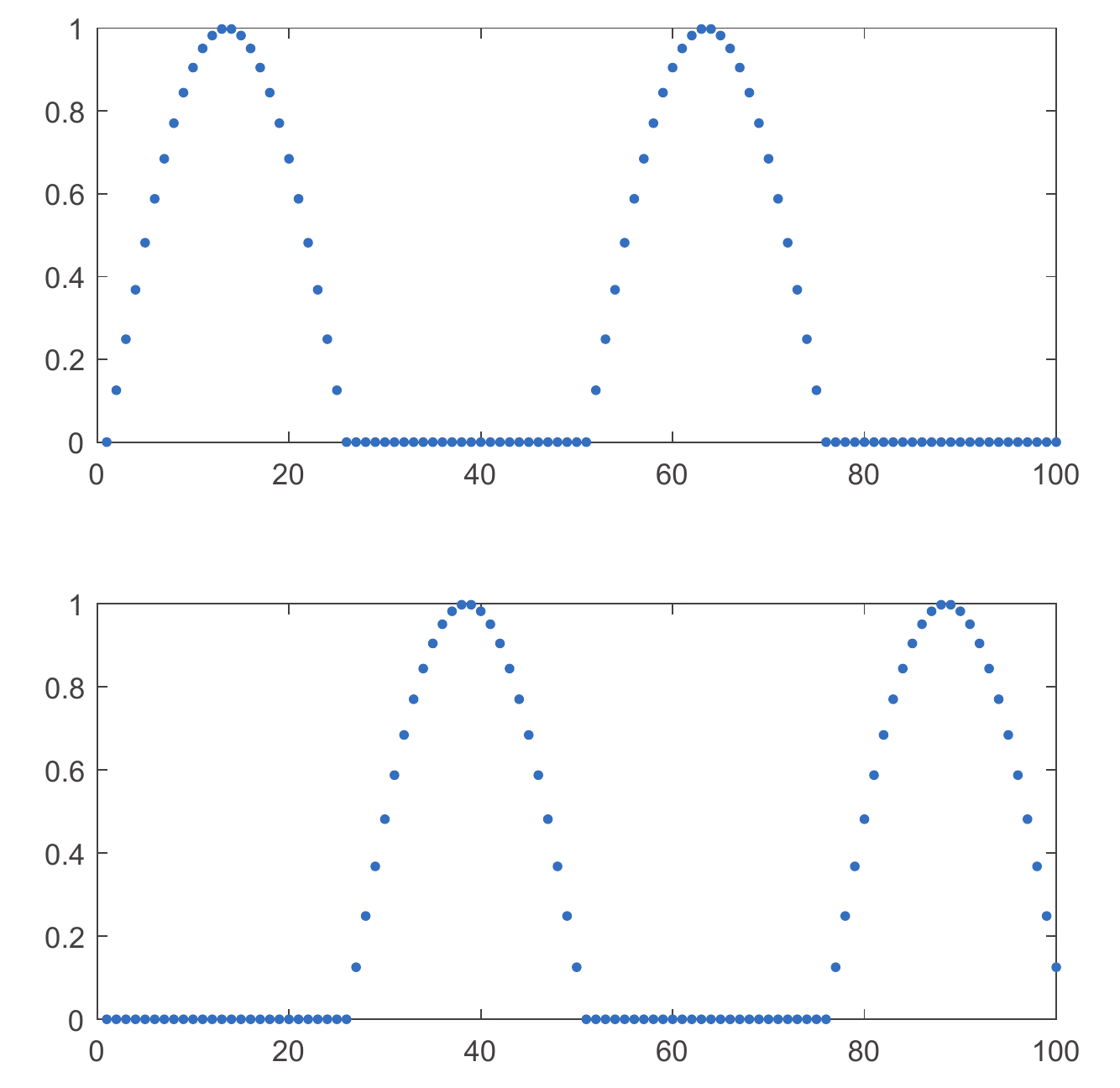} \label{1}
	}
	\quad
	\subfigure[]{
		\includegraphics[scale=0.4]{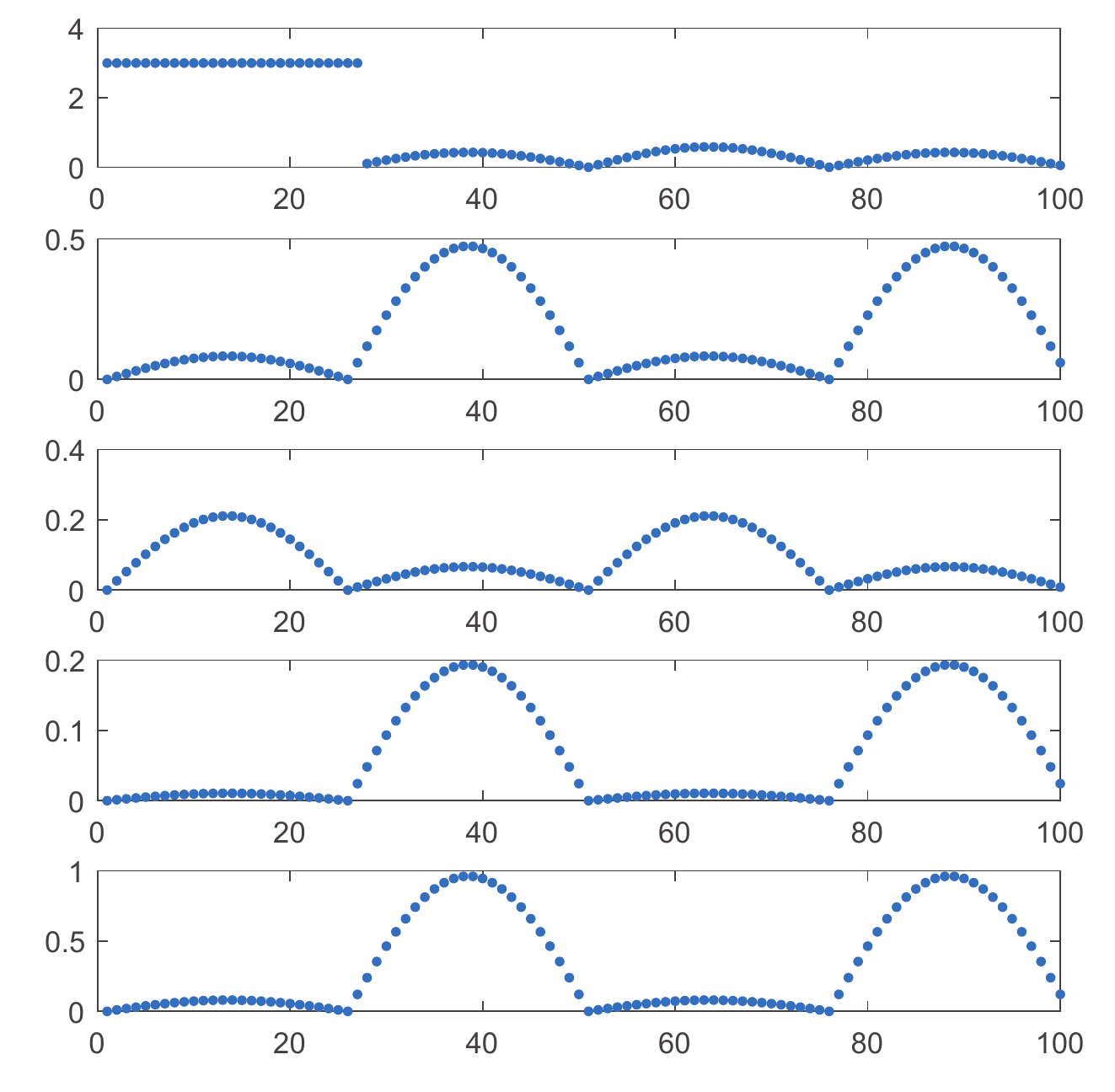} \label{2}
	}
	\quad
	\subfigure[]{
		\includegraphics[scale=0.4]{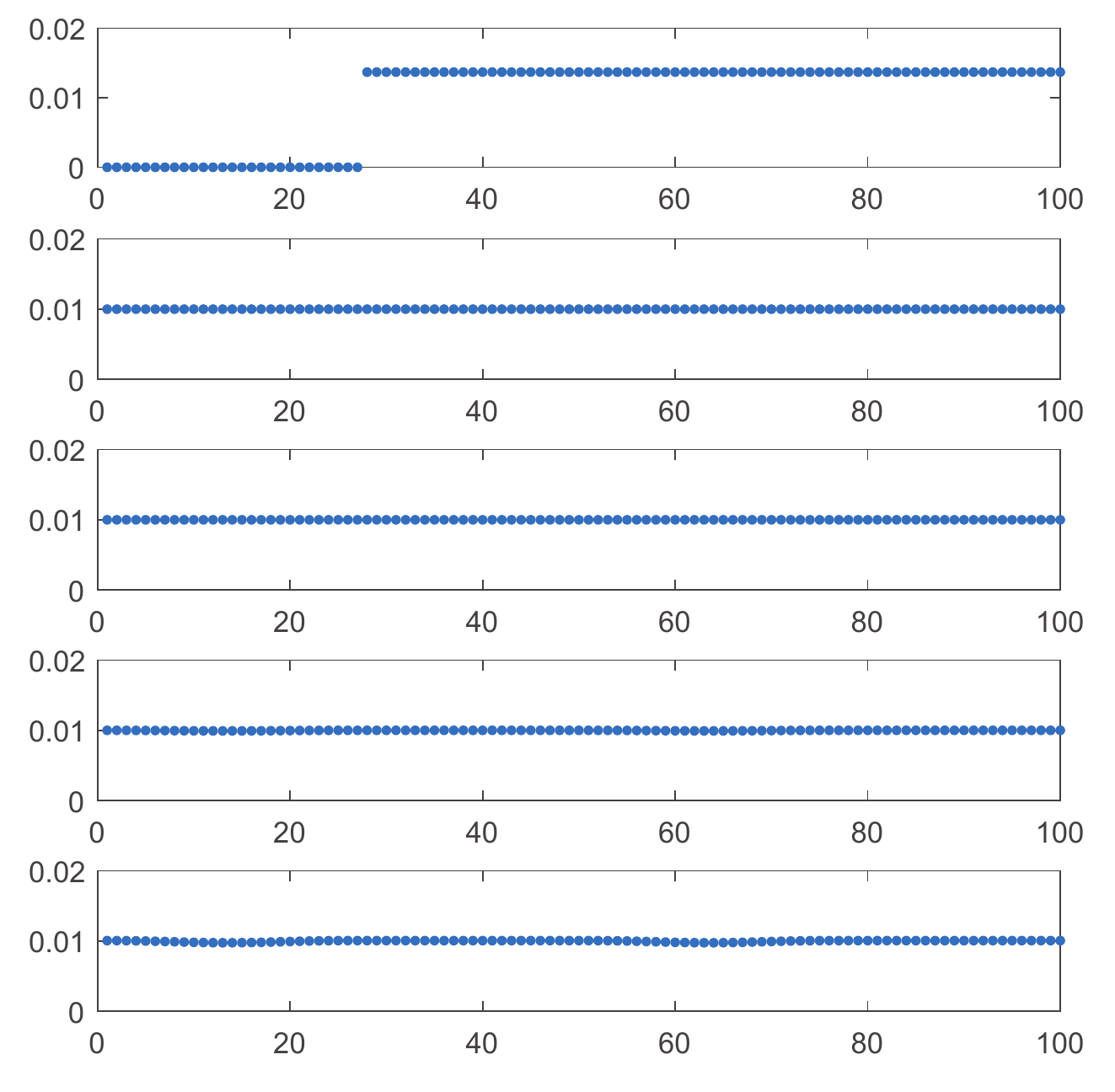}\label{3}
	}
	
	\caption{Illustrative figures of EWNMF: (a) the source signals, (b) the mixed signals, in which the first one is corrupted at the beginning part and (c) the obtained weights corresponding to (b).}
	\label{fig2}
\end{figure*}
\section{experiments}
\subsection{Experimental description}
\indent Experiments were performed on a HP Compaq PC with a 3.40-GHz Core i7-6700 CPU and 16 GB memory, and all the methods were implemented in MATLAB. We compare the performance of the proposed methods with NMF \cite{ref8}, ONMF \cite{ref16}, Semi-NMF \cite{ref13} and Convex-NMF \cite{ref13} on five public data sets, including the Yale, UMISTface, Caltech101, GTFD and TDT2 data sets. \\
\indent The Yale face data set \cite{ref25} was created by Yale University and contains data describing 15 people. Each person has 11 face images with different expressions, postures and lighting. The data set has a total of 150 images, and each has a size of $32\times32$. \\
\indent The UMISTface data set \cite{ref26} was established by the University of Manchester, UK. The data set has a total of 1012 images, including 20 people, each with different angles and different poses. In this experiment, the pixel number of each image is $32\times32$.\\
\indent The Caltech101 data set \cite{ref27} contains 9144 images split between 101 different object categories, as well as an additional background/clutter category. This data set has approximately 40 to 800 images per category, and most categories have approximately 50 images. The size of each image is $32\times32$. \\
\indent The GTFD \cite{ref28} contains 750 images taken in two different sessions and includes 50 people. All people in the data set are represented with cluttered backgrounds. The images show faces with different facial expressions and lighting conditions, and each has a size of $32\times32$. \\
\indent The TDT2 Audio Corpus \cite{ref29} contains six sources, including two news special lines (APW and NYT), two radio programs (VOA and PRI) and two TV programs (CNN and ABC). In this experiment, only the largest 30 categories are used (a total of 9394 documents).\\
\indent Important details of these data sets are shown in Table \ref{tab1}.\\
\begin{table}[!t] \caption{Statistics of the used standard data sets.}
 \label{tab1}	\centering
 \begin{tabular}{lcccccc}  
\toprule   
  Data set & Yale &UMISTface & Caltech101 & GTFD & TDT2 \\  
\midrule   
  Points&165&1012&9144&750&9394   \\  
   Dimensions&1024&1024&1024&1024&36771  \\    
Class&15&20&101&50&30  \\
  \bottomrule  
\end{tabular}
\end{table}
\begin{figure*}[htbp]
	\centering
	\includegraphics[scale=0.45]{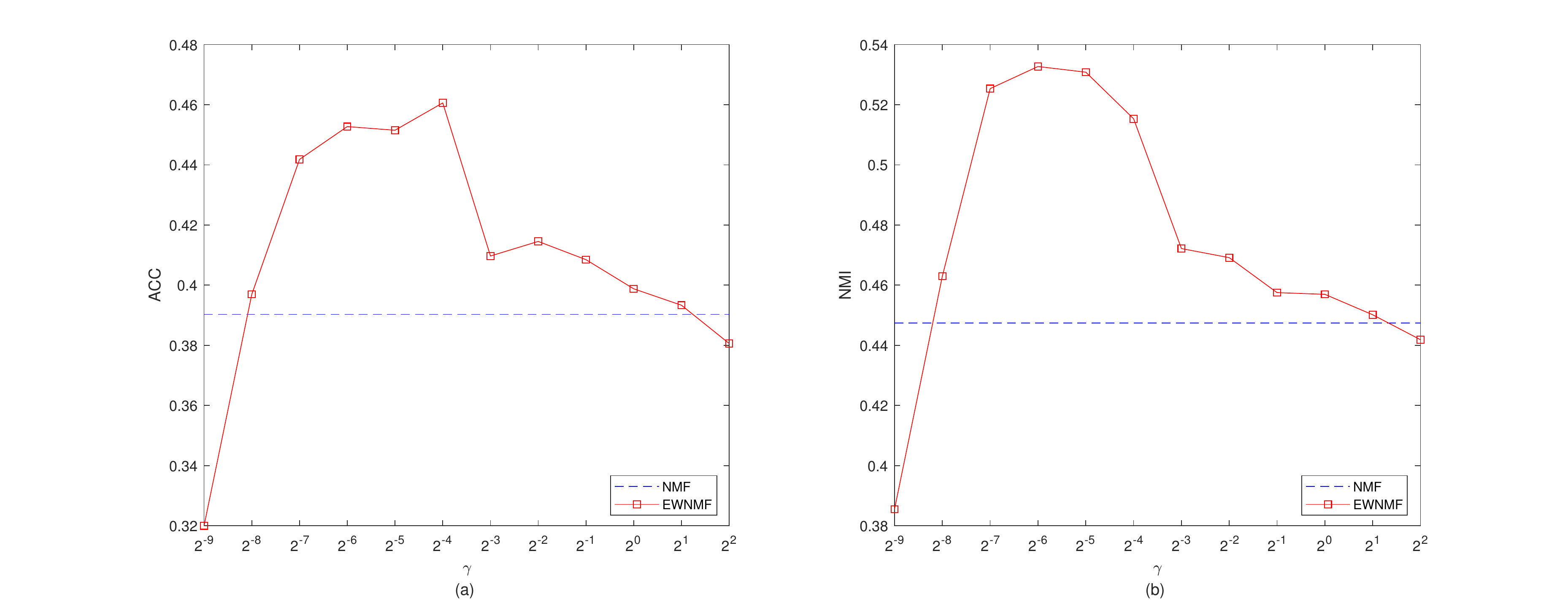}
	\caption{Clustering performance versus the hyperparameter $\gamma$ on the Yale data set: (a) ACC and (b) NMI.}
	\label{fig3}
\end{figure*}
\begin{figure*}[htbp]
	\centering
	\includegraphics[scale=0.45]{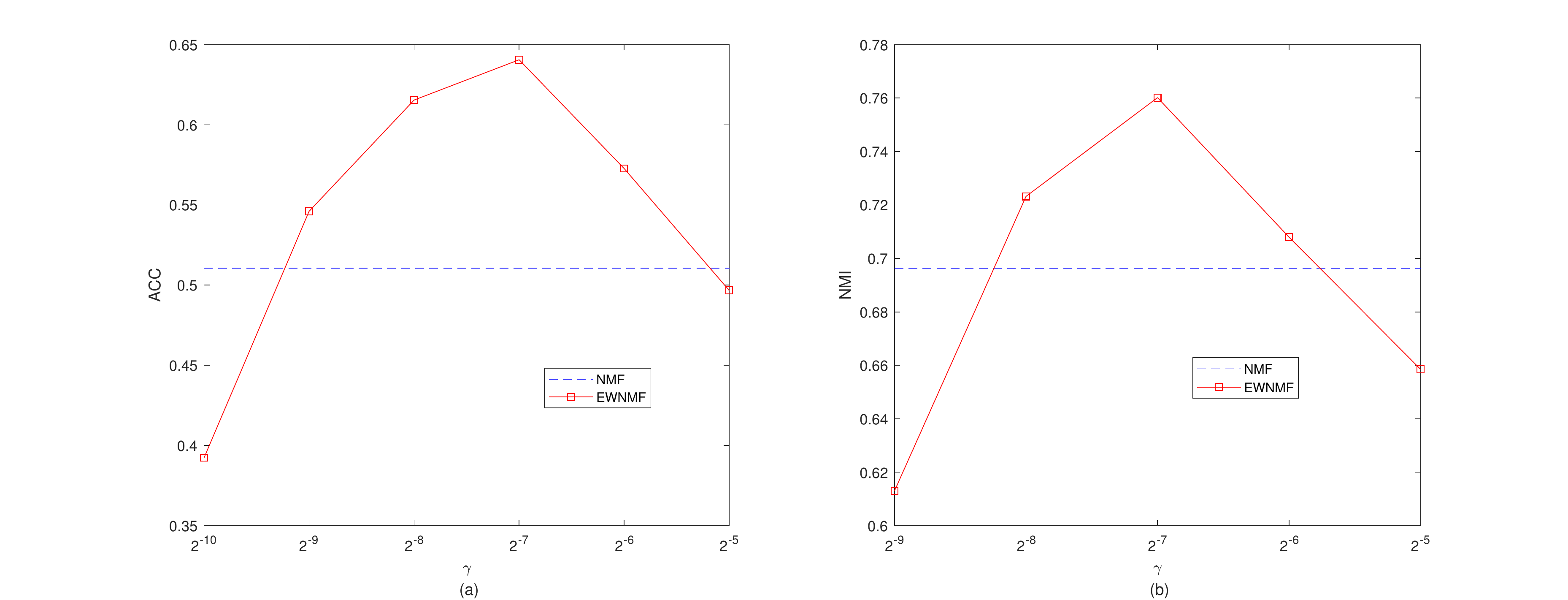}
	\caption{Clustering performance versus the hyperparameter $\gamma$ on the UMISTface data set: (a) ACC and (b) NMI.}
	\label{fig4}
\end{figure*}
\indent After obtaining a new feature representation, we use k-means to cluster them and then compare it with the label to evaluate the clustering results. Clustering accuracy (ACC) \cite{ref30}, \cite{ref31} and normalized mutual information (NMI) \cite{ref32}, \cite{ref33} are used to evaluate the performance of these clustering results.\\
\indent Given a set of the ground true class labels $y$ and the obtained cluster labels $y'$, the clustering accuracy is defined as:\\
\begin{equation}
	ACC=\frac{\sum_{i=1}^N\delta(y_i,map(y'_i))}{N}
	\label{eq26}
\end{equation}
where:\\
$$\delta(a,b)=
\begin{cases}
	1& \text{\emph{a = b}}\\
	0& \text{otherwise}
\end{cases}$$
and $map(\cdot)$ is a permutation mapping function that maps the obtained cluster labels to the real labels. The higher the ACC value is, the better the clustering performance.\\
\indent NMI is used to calculate the agreement between the two data distributions and is defined as follows:\\
\begin{equation}
	NMI(y,y')=\frac{MI(y,y')}{max(H(y),H(y'))}
	\label{eq27}
\end{equation}
where $H(y)$ is the entropy of $y$. $MI(y,y')$ quantifies the amount of information between two random variables (i.e., $y$ and $y'$) and is defined as:
\begin{equation}
	MI(y,y')=\sum_{y_i\in y,y'_j\in y'}p(y_i,y'_j)log(\frac{p(y_i,y'_j)}{p(y_i)p(y'_j)})
	\label{eq28}
\end{equation}
where $p(y_i)$ and $p(y'_j)$ are the probabilities that a data point selected from the data set belongs to the clusters $y_i$ and $y'_j$, respectively; and $p(y_i,y'_j)$ is the joint probability that an arbitrarily selected data point belongs to clusters $y_i$ and $y'_j$ concurrently. The NMI score ranges from 0 to 1, and the larger NMI is, the better the clustering performance. 

\begin{table}[!t] \caption{ACC and NMI on the Yale data set (\%).}
	\label{tab2}	\centering
	\begin{tabular}{lccccc}  
		\toprule   
		Method & NMF & ONMF & Semi-NMF & Convex-NMF & EWNMF \\  
		\midrule   
		ACC& 39.03 & 41.21 & 42.48 & 31.88 & \bf{46.06}   \\  
		NMI& 44.74 & 47.36 & 48.48 & 37.91 & \bf{53.27}   \\    
		\bottomrule  
	\end{tabular}
\end{table}
\begin{table}[!t] \caption{ACC and NMI on the UMISTface data set (\%).}
	\label{tab3}	\centering
	\begin{tabular}{lccccc}  
		\toprule   
		Method & NMF & ONMF & Semi-NMF & Convex-NMF & EWNMF \\  
		\midrule   
		ACC& 51.06 & 50.99 & 52.15 & 23.52 & \bf{64.05}   \\  
		NMI& 69.63 & 68.34 & 67.70 & 29.13 & \bf{76.01}   \\    
		\bottomrule  
	\end{tabular}
\end{table}
\begin{figure*}[htbp]
	\centering
	\includegraphics[scale=0.45]{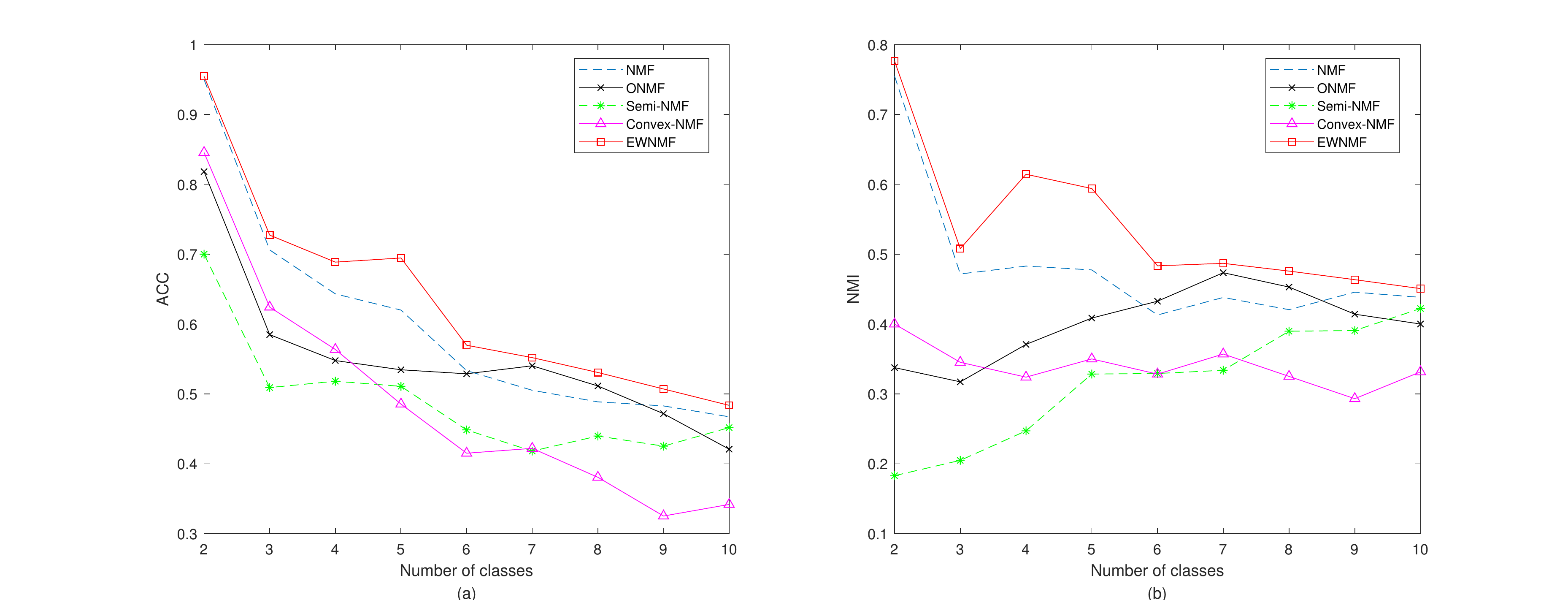}
	\caption{Clustering performance versus cluster number on the Yale data set: (a) AC and (b) NMI.}
	\label{fig5}
\end{figure*}
\begin{figure*}[h]   
	\centering
	\includegraphics[scale=0.45]{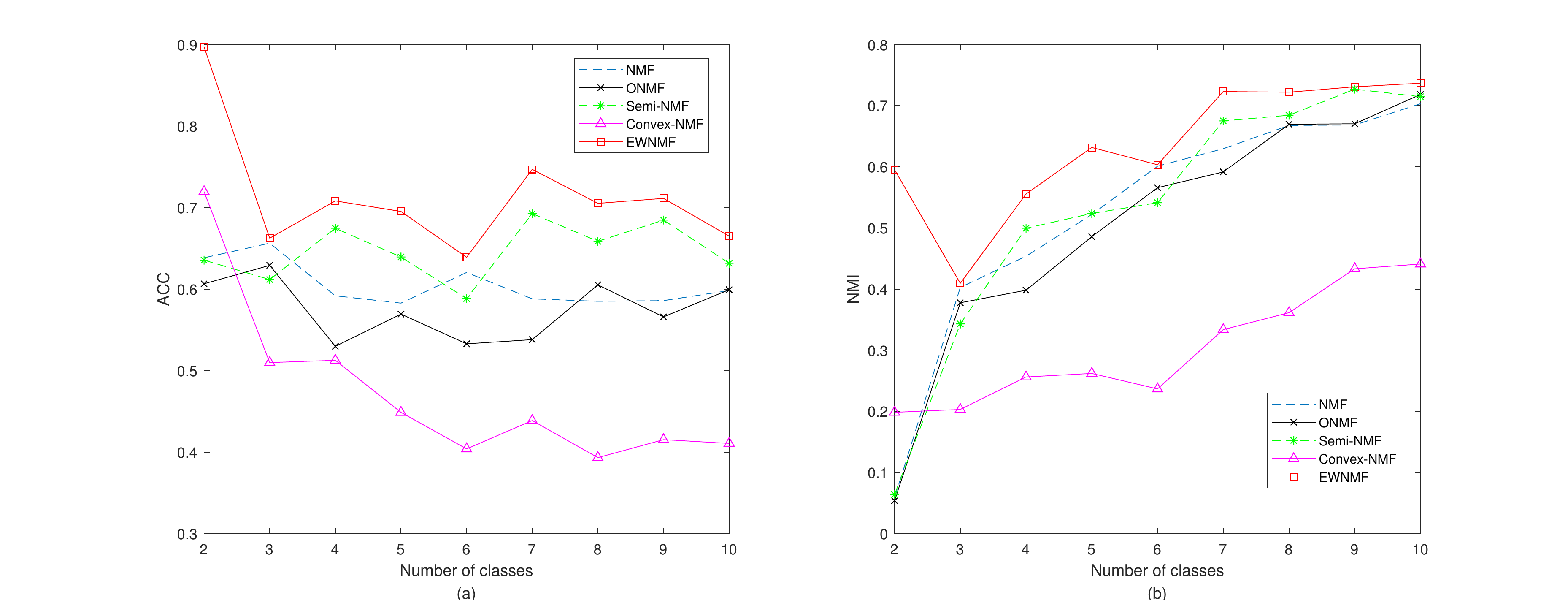}
	\caption{Clustering performance versus cluster number on the UMISTface data set: (a) AC and (b) NMI.}
	\label{fig6}
\end{figure*}
\begin{figure*}[h]   
	\centering
	\includegraphics[scale=0.45]{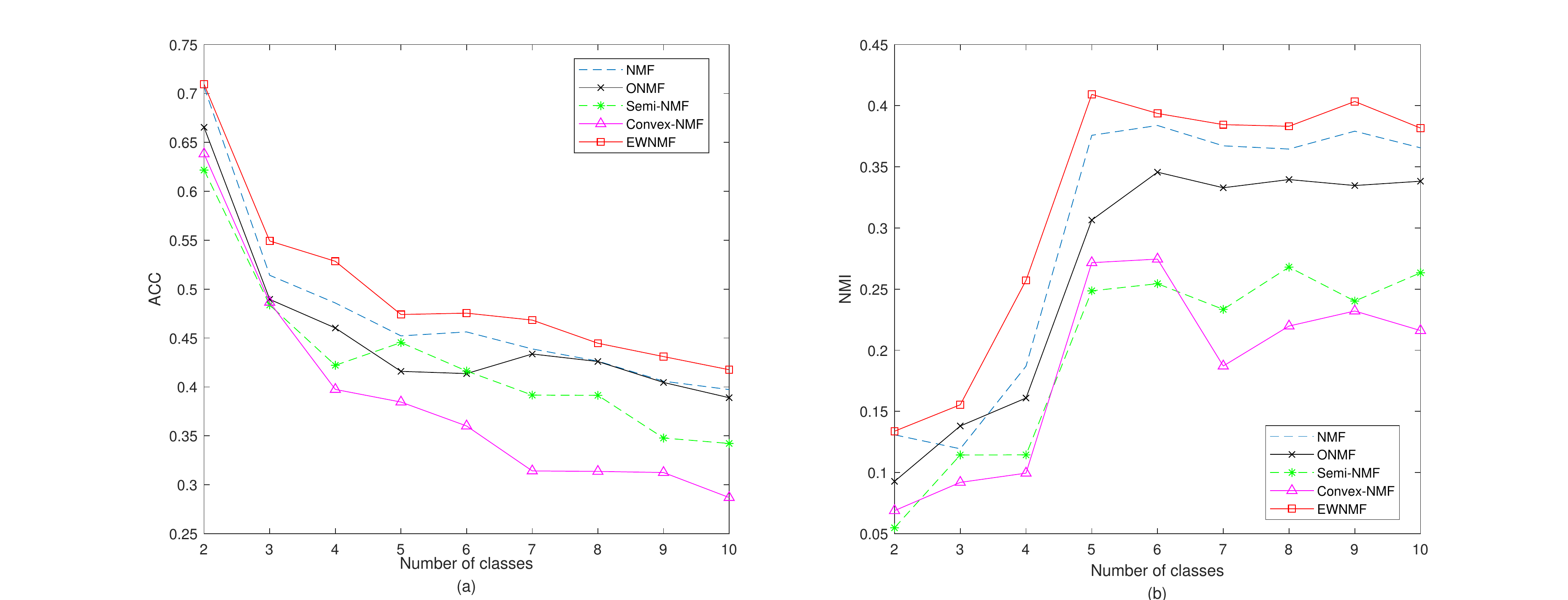}
	\caption{Clustering performance versus cluster number on the Caltech101 data set: (a) AC and (b) NMI.}
	\label{fig7}
\end{figure*}

 Before the experiment, we normalized all the data sets to scale the minimum and maximum values of each data point to 0 and 1, respectively. All the methods use the same random distribution for initialization of $W$ and $H$ to make them uniformly distributed on [0.1 1.1] and perform 300 iterations to ensure sufficient convergence.\\
\subsection{Experimental results}
\subsubsection{Signal unmixing on synthetic data} We use a synthetic data set to simulate a signal-mixture process, which mixes two source signals by a uniformly distributed matrix on [0 1] and then destroys the beginning part of the first one. The destroyed part in the mixed signal will hinder signal recovery. Then, we use EWNMF to unmix the signals to demonstrate the usefulness of the proposed entropy weighted method. Fig. \ref{fig2} shows the source signals, mixed signals (including the destroyed signal) and the obtained weights. The weights of the destroyed elements in the first signal are small and can even be considered as zero, and the other weights are similar. Thus, we can conclude that EWNMF can provide correct weights to the importance of attributes in the data set.
\subsubsection{EWNMF compared with the standard NMF} We investigate the ability of the entropy weighted strategy to improve the performance of the standard NMF. We apply NMF and EWNMF to reduce the dimension number of the Yale and UMISTface data sets and then use k-means to cluster these new representations. Note that the number of reduced dimensions is equal to that of the clusters. Figs. \ref{fig3} and \ref{fig4} show the clustering results evaluated by ACC and NMI. EWNMF indeed provides better performance than the standard NMF and can achieve a consistently superior performance to the standard NMF on a wide range of the hyperparameter $\gamma$, demonstrating the robustness of EWNMF.\\
\indent The optimal clustering results on the entire Yale and UMISTface data sets are shown in Tabs. \ref{tab2} and \ref{tab3}. An interesting observation is highlighted and shows a consistent result with the above figures that EWNMF still provides the best evaluation standards.
\begin{figure*}[h]   
	\centering
	\includegraphics[scale=0.45]{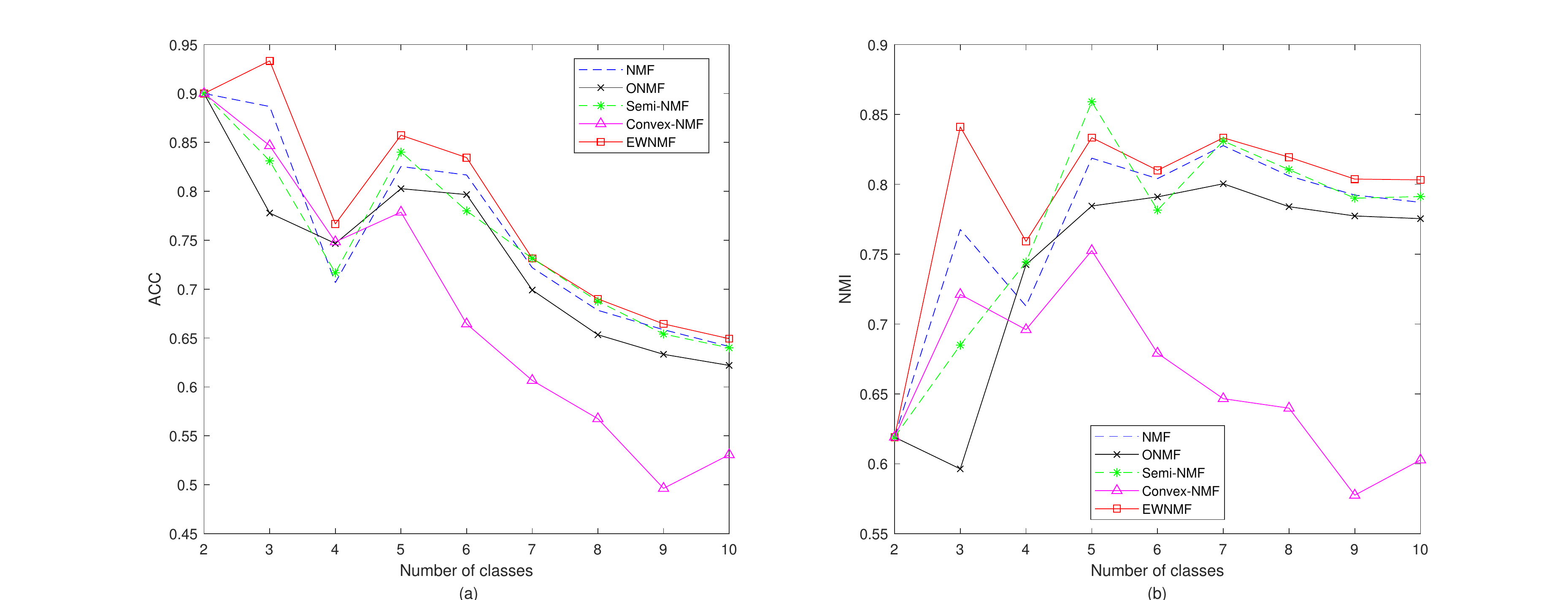}
	\caption{Clustering performance versus cluster number on the GTFD data set: (a) AC and (b) NMI.}
	\label{fig8}
\end{figure*}
\begin{figure*}[h]   
	\centering
	\includegraphics[scale=0.45]{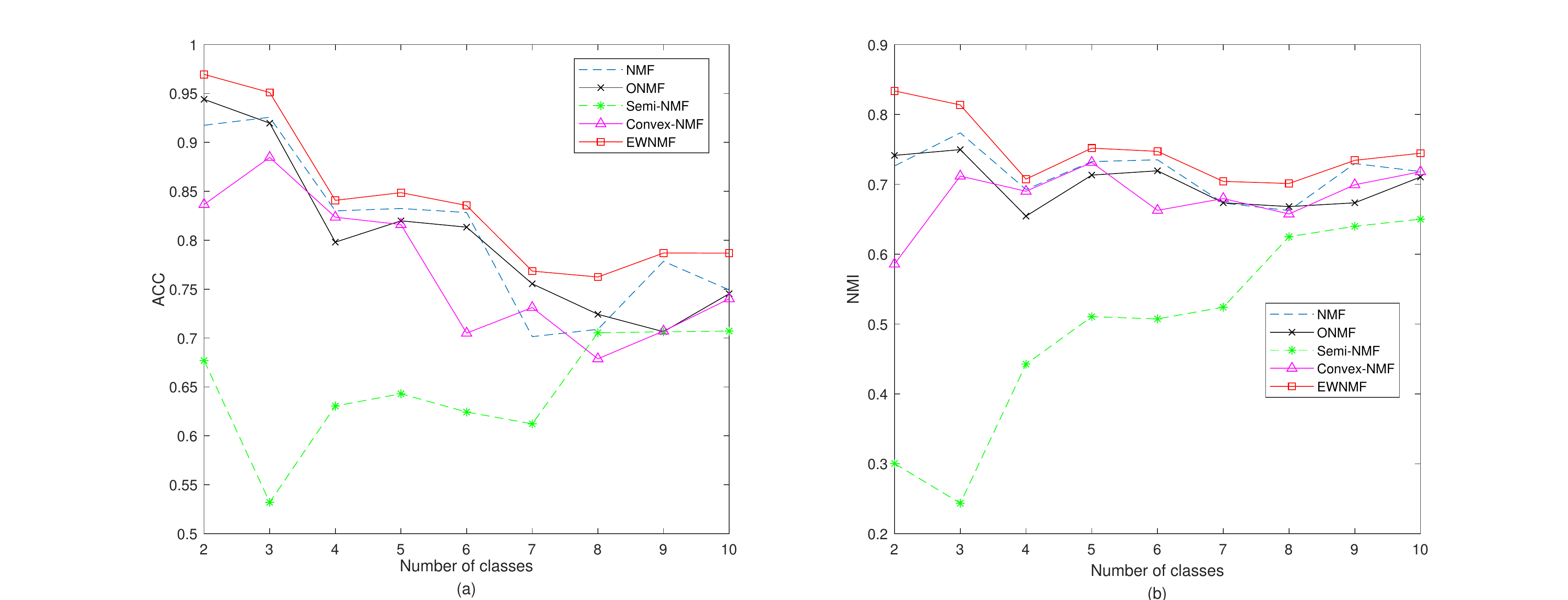}
	\caption{Clustering performance versus cluster number on the TDT2 data set: (a) AC and (b) NMI.}
	\label{fig9}
\end{figure*}
\begin{figure*}[htbp]
	\centering
	
	\subfigure[]{
		\includegraphics[scale=0.35]{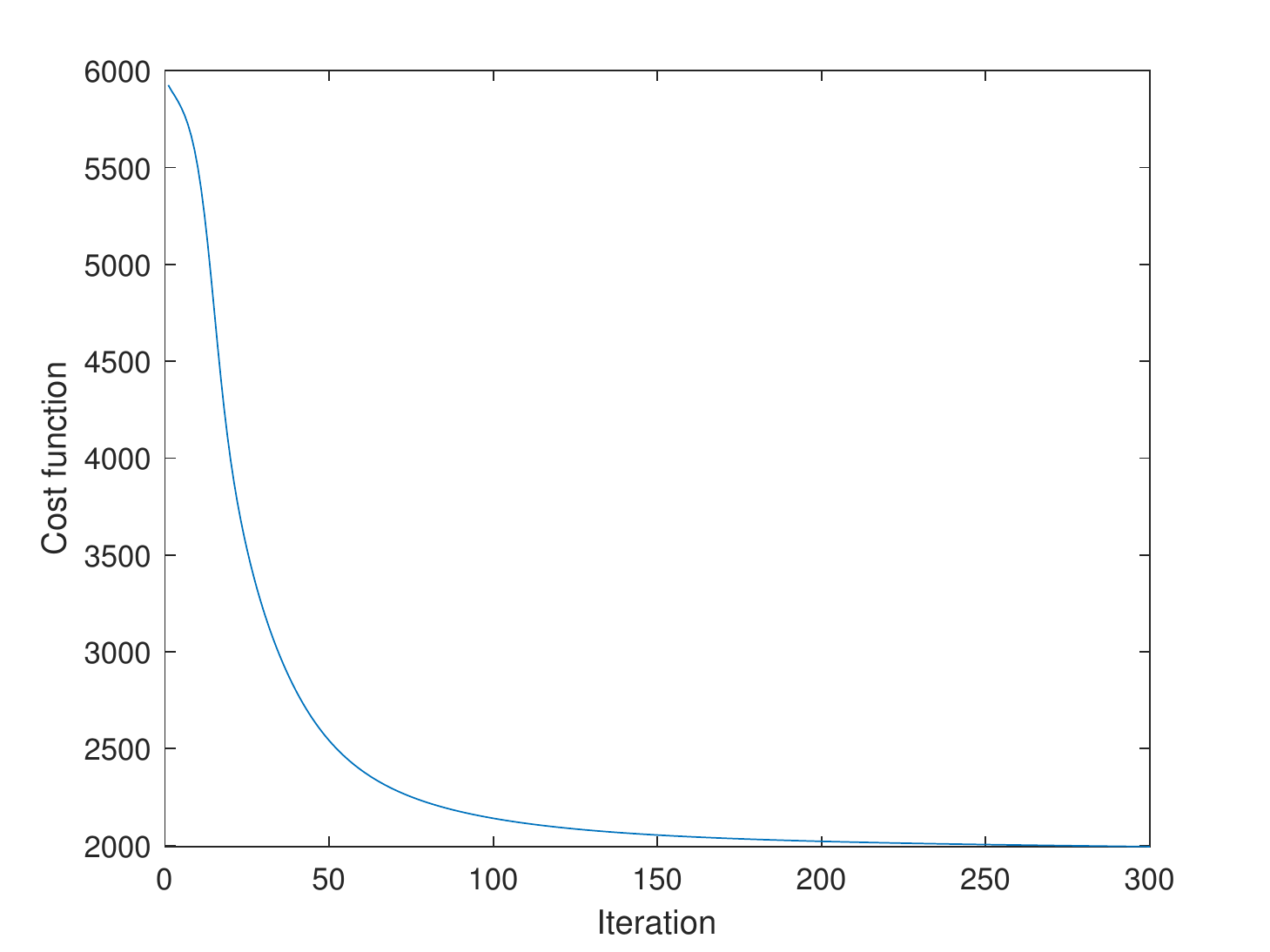} \label{1}
	}
	\quad
	\subfigure[]{
		\includegraphics[scale=0.35]{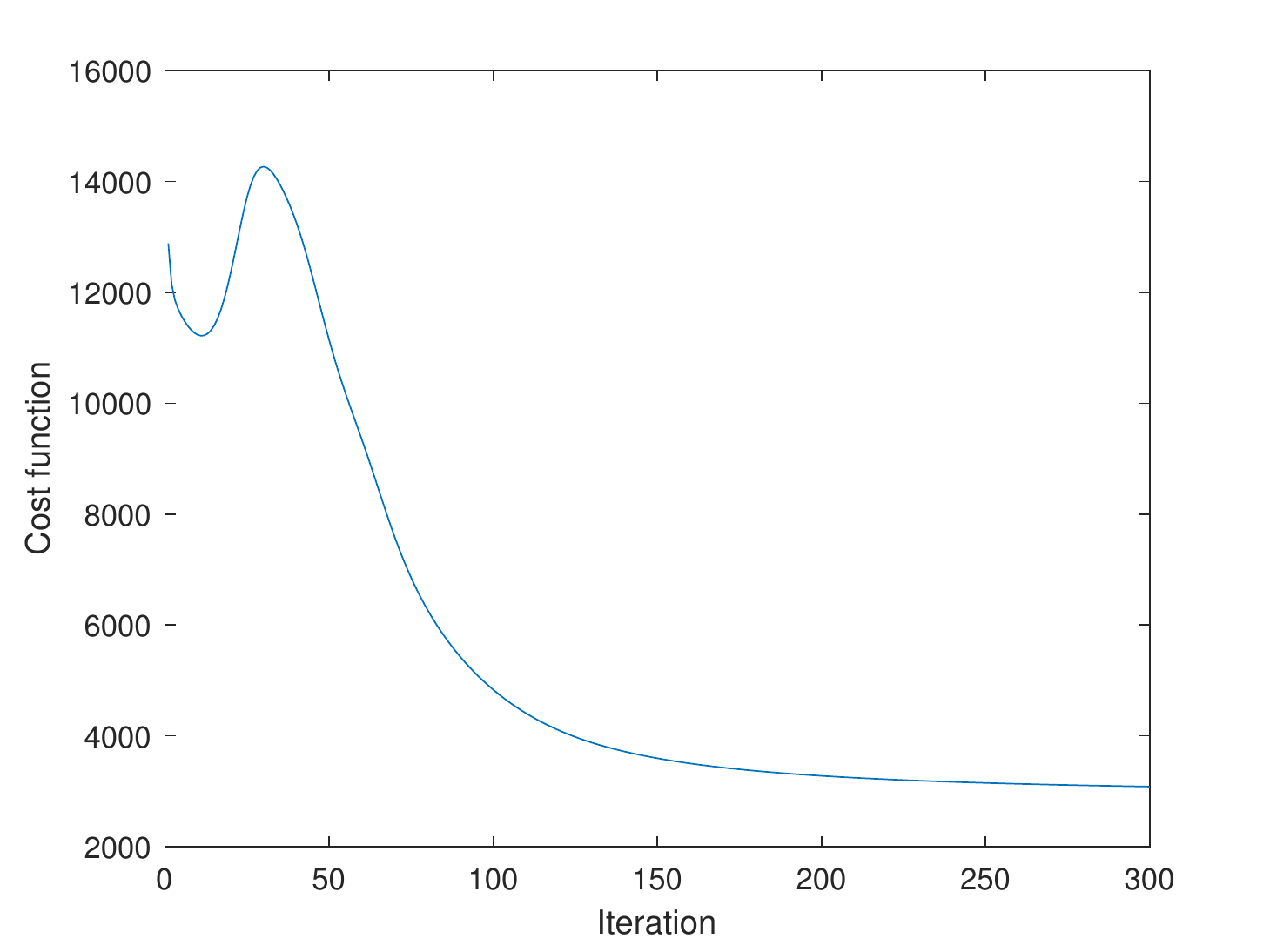} \label{2}
	}
	
	\quad
	\subfigure[]{
		\includegraphics[scale=0.35]{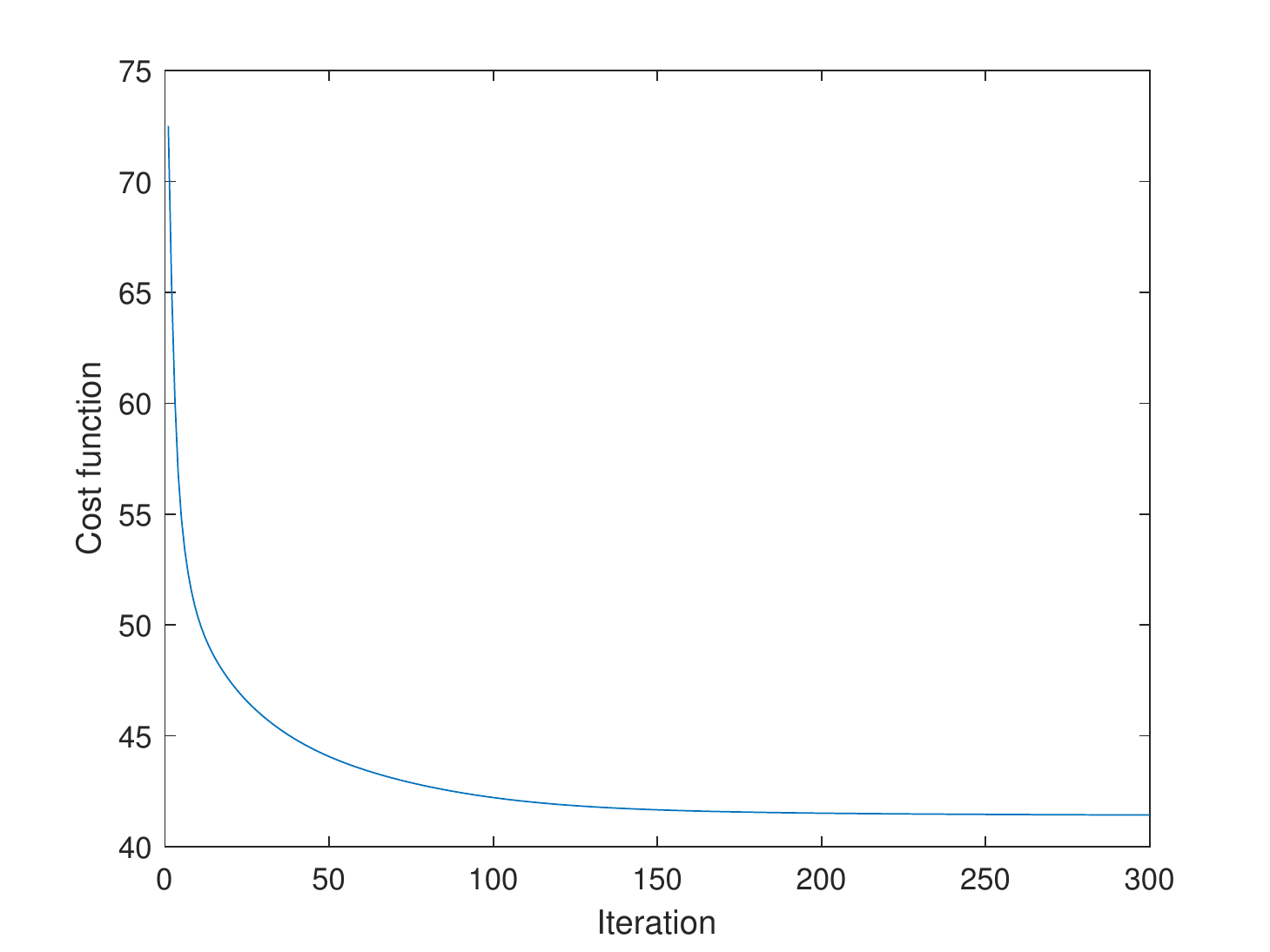}\label{3}
	}
	\quad
	\subfigure[]{
		\includegraphics[scale=0.35]{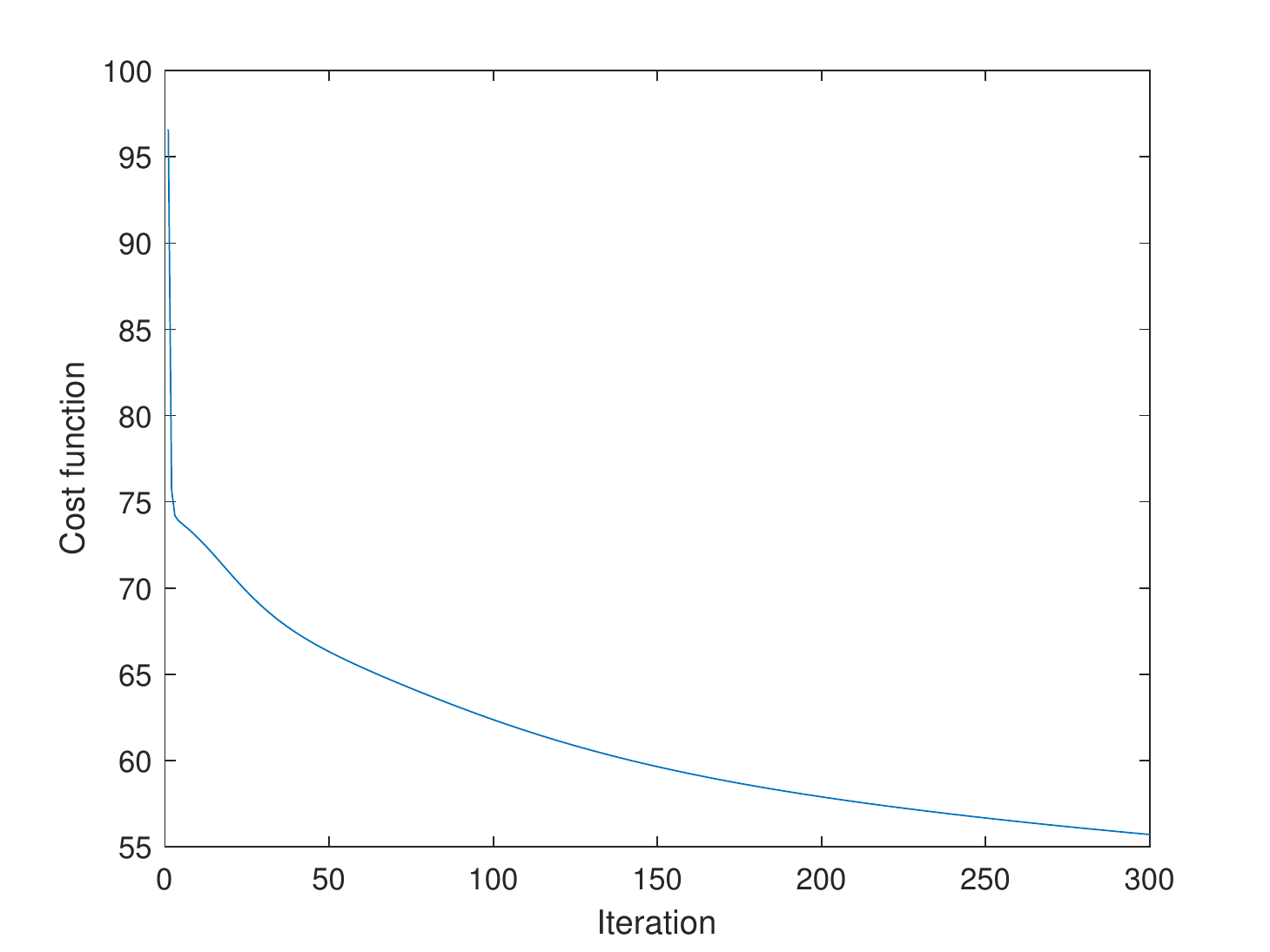}\label{4}
	}
	\quad
	\subfigure[]{
		\includegraphics[scale=0.35]{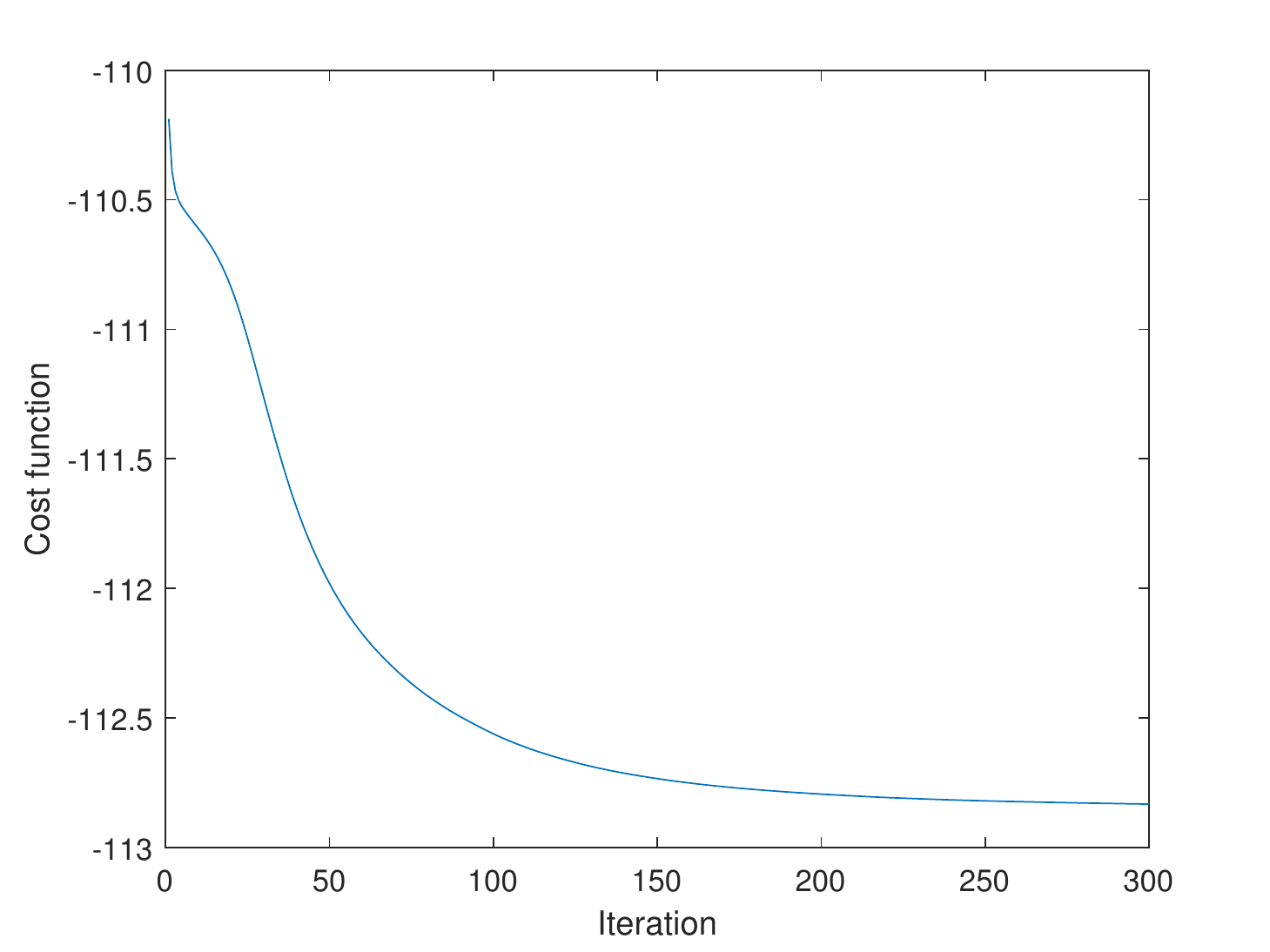}\label{5}
	}
	
	\caption{Cost function on the Yale data set. From left to right, they are obtained by (a) NMF, (b) ONMF, (c) Semi-NMF, (d) Convex-NMF and (e) EWNMF.}
	\label{fig10}
\end{figure*}
\subsubsection{Clustering results with different cluster numbers} We study the relationship between the evaluation standards and cluster number. The hyperparameter $\gamma$ is selected in $\{10^i, i=-8, -7, ..., 7, 8\}$ to obtain the optimal results in a large range. Because the NMF problem does not have a sole solution, we randomly initialized 10 times to obtain a credible averaged ACC and NMI. Different cluster numbers ranging from 2 to 10 are selected. In a certain data set with \emph{k} clusters, the experimental details are described as follows:\\
\indent 1) Randomly select \emph{k} categories as a subset for the following experiment.\\
\indent 2) Randomly initialize $W$ and $H$, obtain new representations, and cluster them by k-means. Note that EWNMF uses the selected hyperparameter $\gamma$ according to the above instruction.\\
\indent 3) Repeat 1) and 2) 10 times to obtain an average result.\\
\begin{figure*}[htbp]
	\centering
	
	\subfigure[]{
		\includegraphics[scale=0.45]{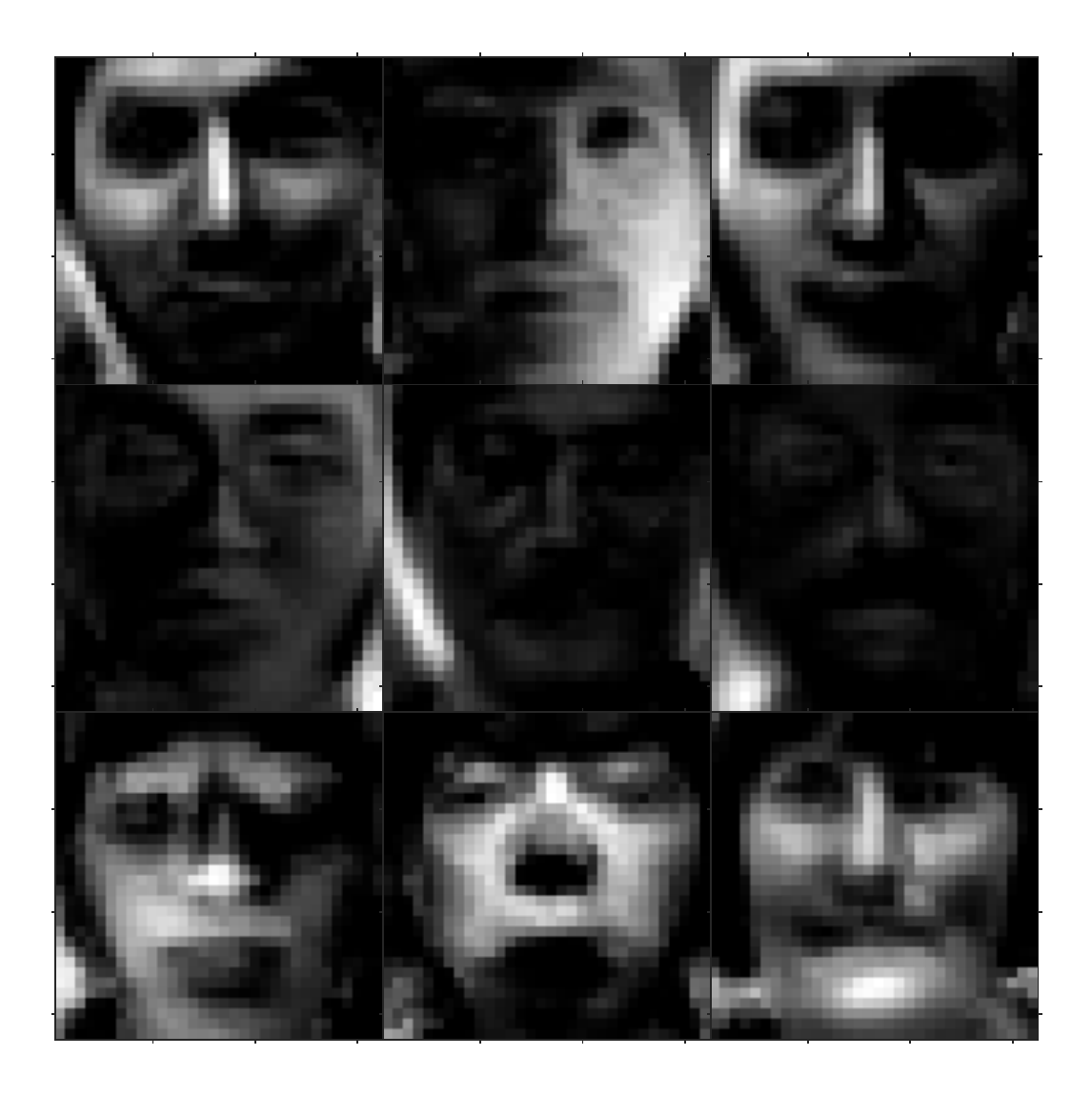} \label{1}
	}
	\quad
	\subfigure[]{
		\includegraphics[scale=0.45]{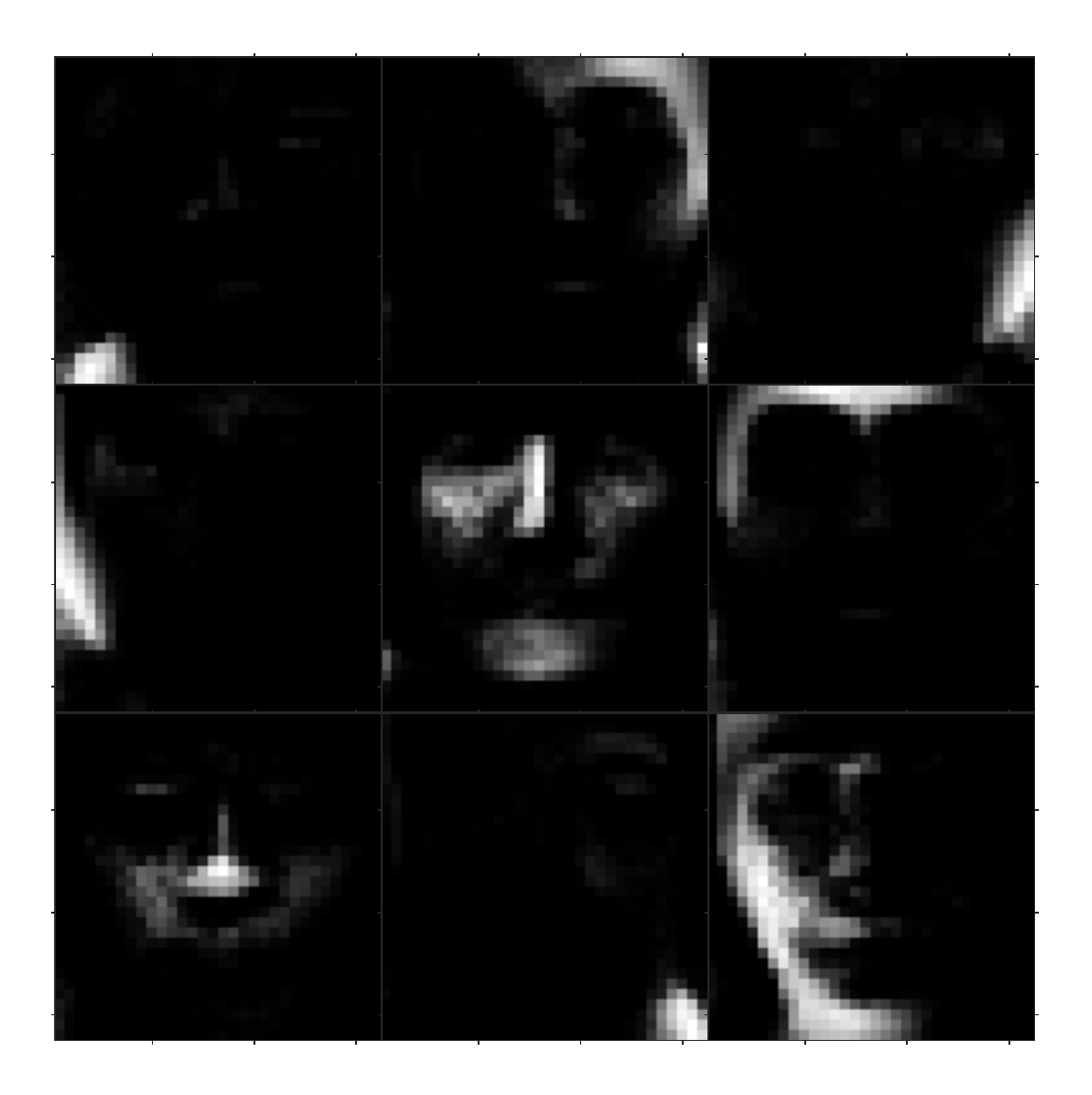} \label{2}
	}	
	
	\quad
	\subfigure[]{
		\includegraphics[scale=0.45]{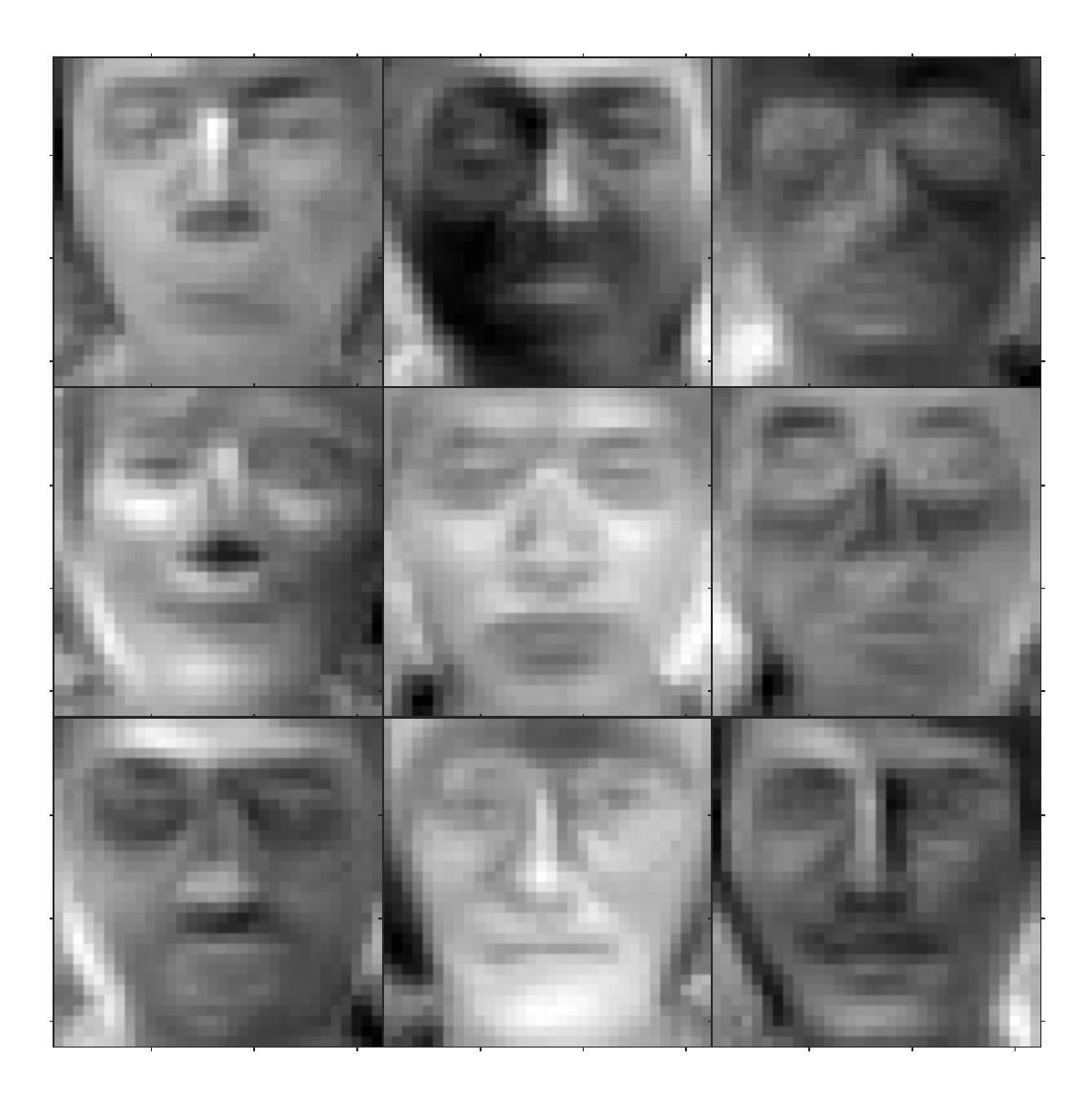}\label{3}
	}
	\quad
	\subfigure[]{
		\includegraphics[scale=0.45]{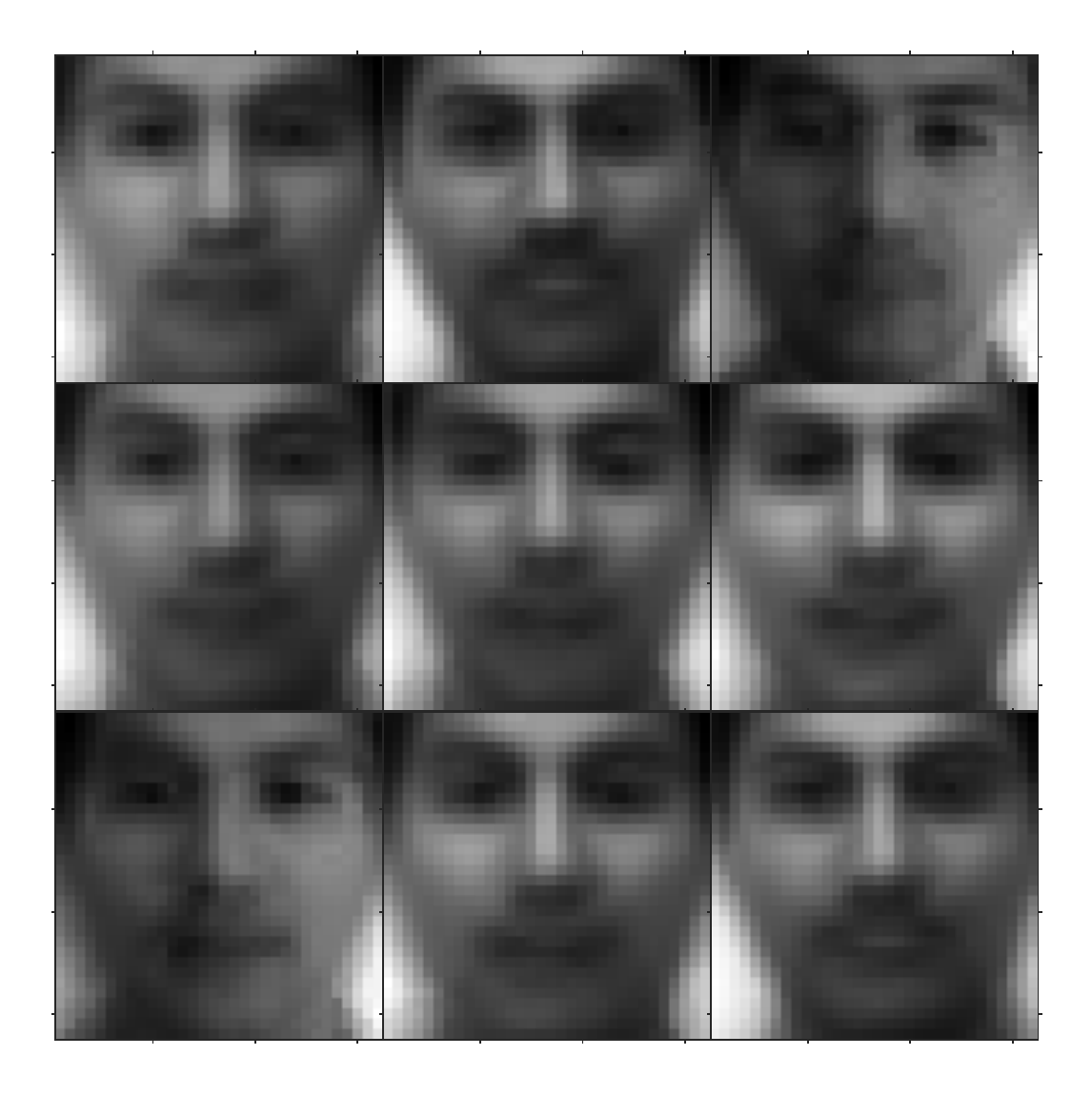}\label{4}
	}
	\quad
	\subfigure[]{
		\includegraphics[scale=0.45]{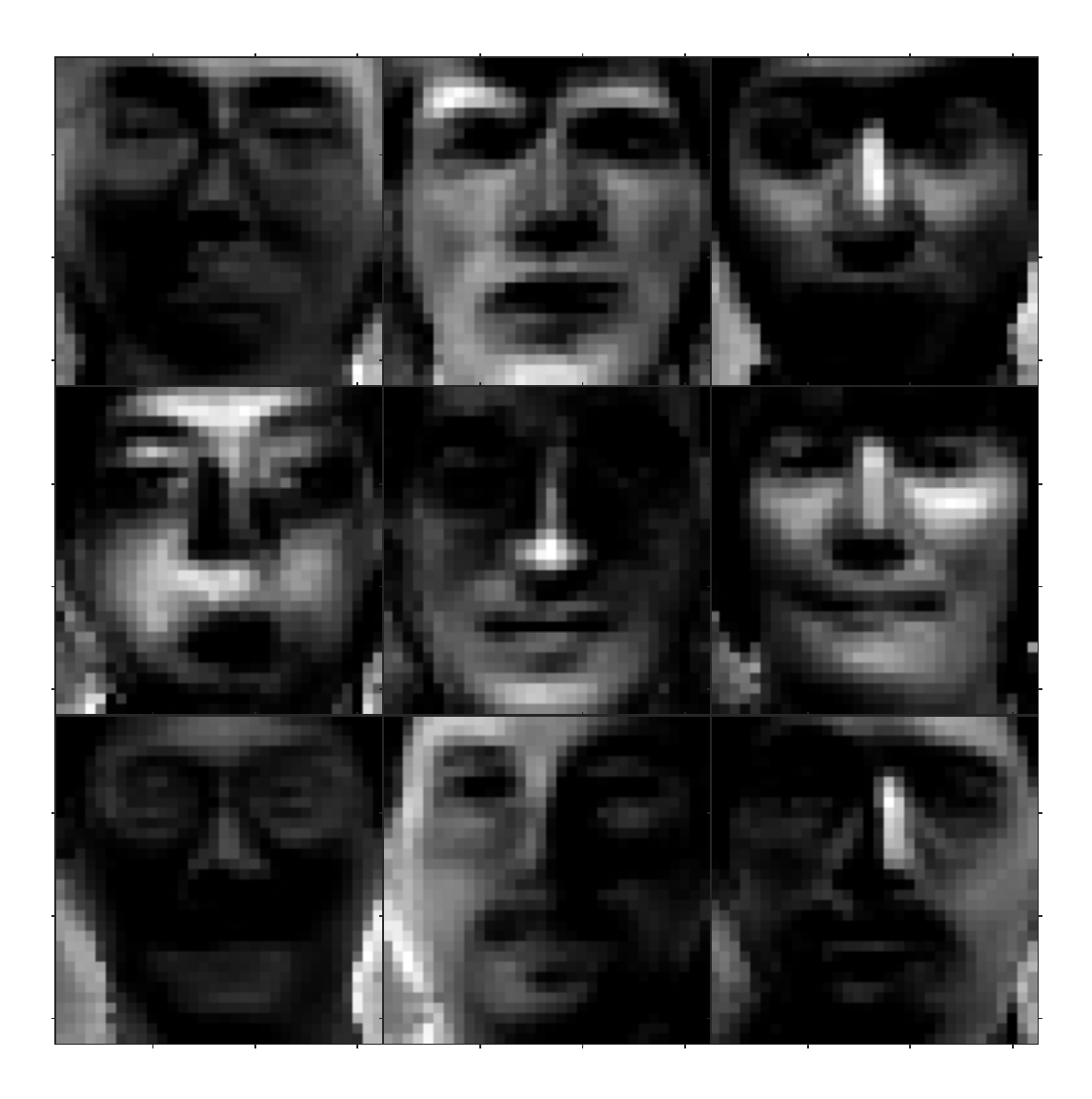}\label{5}
	}
	
	\caption{Some base images with respect to the Yale data set, in which the summation of every column is normalized to one. From left to right, they are obtained by (a) NMF, (b) ONMF, (c) Semi-NMF, (d) Convex-NMF and (e) EWNMF.}
	\label{fig11}
\end{figure*}
\indent The clustering results, ACC and NMI versus the number of clusters, are reported in Figs. \ref{fig5}-\ref{fig9}. The proposed method can generally yield more accurate clustering results, and only for two clusters of the GTFD data set, Semi-NMF occasionally exceeds the accuracy of the proposed method. Also, none of the other methods performs better in all aspects than EWNMF.
\\
\subsubsection{Convergence speed and part-based learning} From the theoretical analysis described above, we can conclude that the proposed method is monotonically decreasing, but due to the nonconvexity of the cost function, it cannot be guaranteed to be strictly convergent \cite{ref34}. Thus, we investigate the convergence speed of the NMF methods.

Fig. \ref{fig10} shows the cost functions of NMF, ONMF, Semi-NMF, Convex-NMF and EWNMF on the Yale data set. Except for the Convex-NMF method, which converges slowly, the other methods reached a stable point within 200 iterations, demonstrating that they have similar convergence speeds. However, ONMF cannot guarantee the monotonic decrease of the cost function because it uses a proximal Lagrange multiplier method. Fig. \ref{fig11} shows the base images on the Yale data set and demonstrates that Semi-NMF and Convex-NMF identify more global faces. EWNMF has poor locality compared to ONMF; however, the former has a better clustering effect than the latter. It is visually difficult to evaluate the performance of the NMF methods, even though they are all markedly different.
\section{Conclusion}
\indent This paper proposes a new NMF method, which adds weights to each attribute of each data to emphasize their importance. We introduce an entropy regularizer to consider these weights as the probabilities of importance within [0 1], which mimics the process of human reasoning more accurately. These weights can be solved by the Lagrange multiplier method, and a simple update is achieved. The experimental results show that the proposed method produces performance that is competitive with those of existing methods.\\
\indent The entropy regularizer requires an additional hyperparameter to control the certainty of the weights. In the future, we plan to develop an auto-adjustment strategy for this hyperparameter.
\section*{Acknowledgment}
\indent The authors would like to thank the editor and anonymous reviewers for their constructive comments and suggestions.

\begin{IEEEbiography}[{\includegraphics[width=1in,height=1.25in,clip,keepaspectratio]{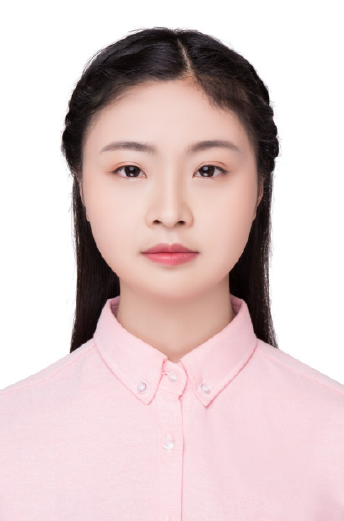}}]{Jiao Wei} is pursuing her M.S. degree in the School of Medical and Bioinformatics Engineering at Northeastern University and received her B.S. degree from Qufu Normal University in 2014. Her research interests include machine learning and nonnegative matrix decomposition algorithms.
\end{IEEEbiography}
\begin{IEEEbiography}[{\includegraphics[width=1in,height=1.25in,clip,keepaspectratio]{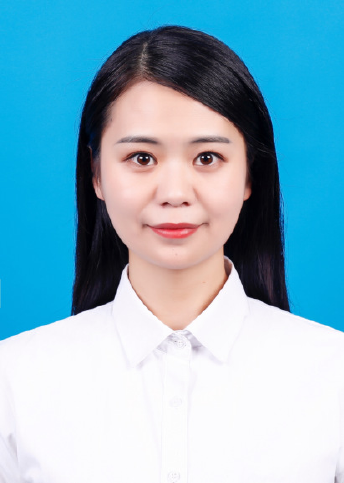}}]{Tong Can} is a Ph.D. student and received a bachelor's degree in mathematics and applied mathematics from Northeast University in 2014 and a master's degree in computational mathematics from Northeastern University in 2018. Her primary research interests include basic algorithm theory and acceleration methods of machine learning.
\end{IEEEbiography}
\begin{IEEEbiography}[{\includegraphics[width=1in,height=1.25in,clip,keepaspectratio]{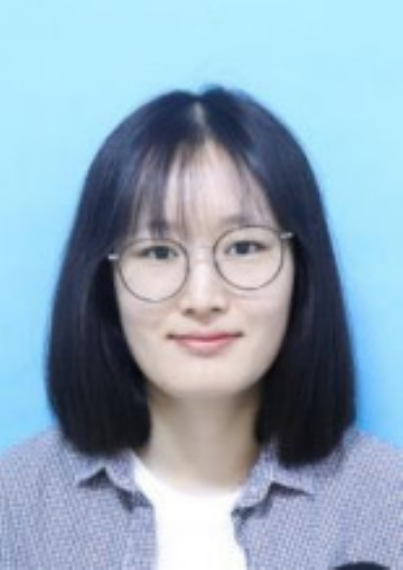}}]{Bingxue Wu} is pursuing her M.S. degree in the School of Medical and Bioinformatics Engineering at Northeastern University and received her B.S. degree from Harbin Business University in 2015. Her research interests are machine learning and compressive perception in medical imaging.
\end{IEEEbiography}
\begin{IEEEbiography}[{\includegraphics[width=1in,height=1.25in,clip,keepaspectratio]{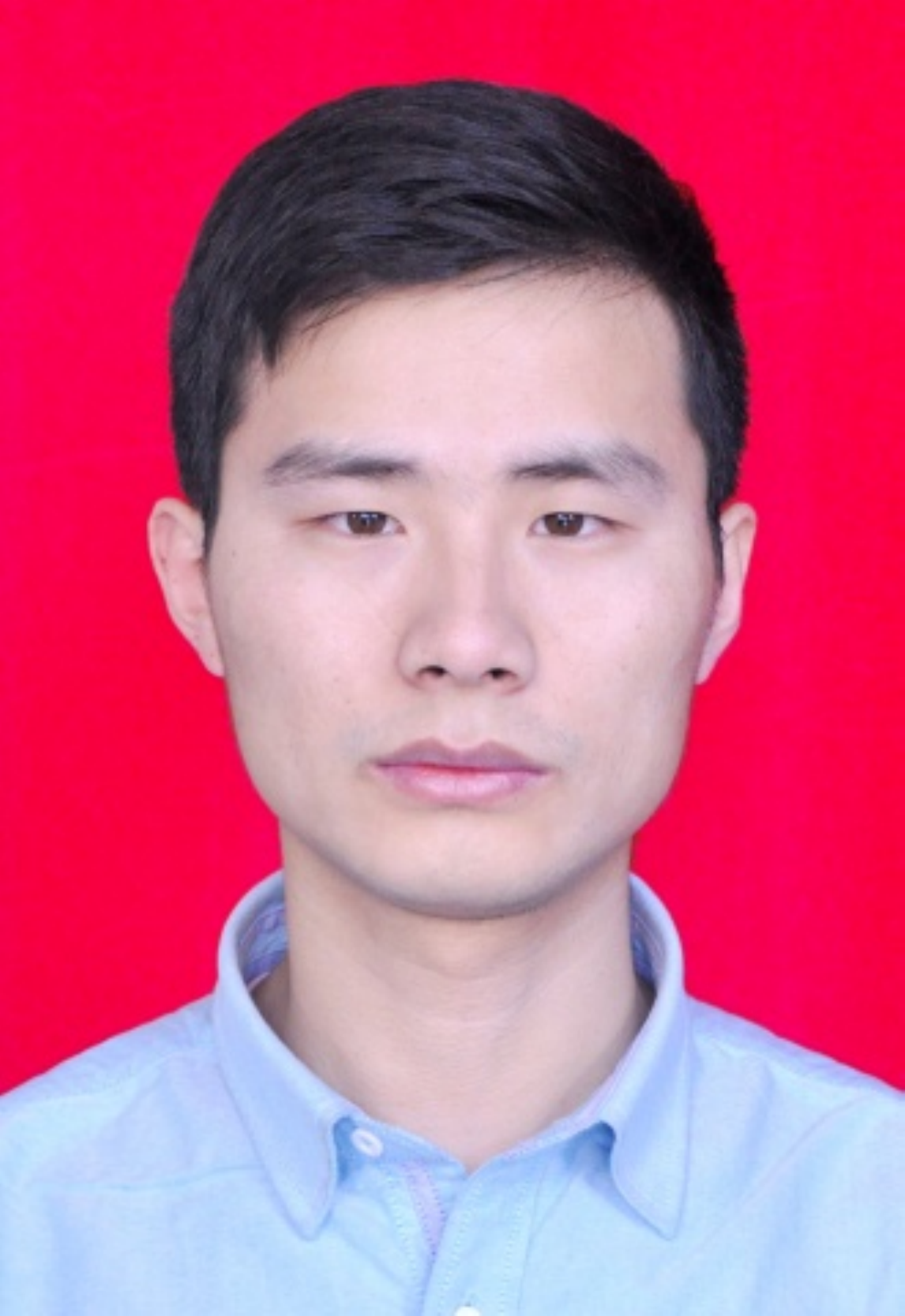}}]{Qiang He} received the Ph.D. degree in computer
	application technology from the Northeastern University, Shenyang, China, in 2020. He is currently an Associate Professor with the College of Medicine and Biological Information Engineering, Northeastern University, Shenyang, China. His research interests include social network analytic, machine learning, etc.
\end{IEEEbiography}
\begin{IEEEbiography}[{\includegraphics[width=1in,height=1.25in,clip,keepaspectratio]{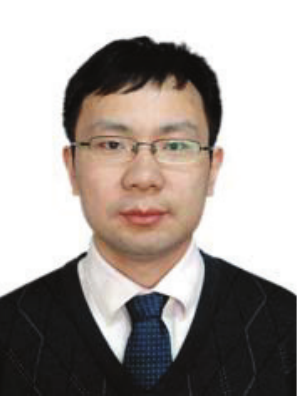}}]{Shouliang Qi} is an associate professor at Northeastern University, China. After he received a Philosophy degree doctor from Shanghai Jiao Tong University in 2007, he joined the GE Global Research Center and was responsible for designing an innovative magnetic resonance imaging (MRI) system. In 2014-2015, he worked as a visiting scholar at Eindhoven University of Technology and Epilepsy Center Kempenhaeghe, the Netherlands. In recent years, he has been conducting productive studies in advanced MRI technology, intelligent medical imaging computing and modeling, and the convergence of Nano-Bio-Info-Cogn (NBIC). He has published more than 60 papers in peer-reviewed journals and international conferences. He has won many academic awards, such as Chinese Excellent PH. D. Dissertation Nomination Award, and the Award for Outstanding Achievement in Scientific Research from Ministry of Education.

\end{IEEEbiography}
\begin{IEEEbiography}[{\includegraphics[width=1in,height=1.25in,clip,keepaspectratio]{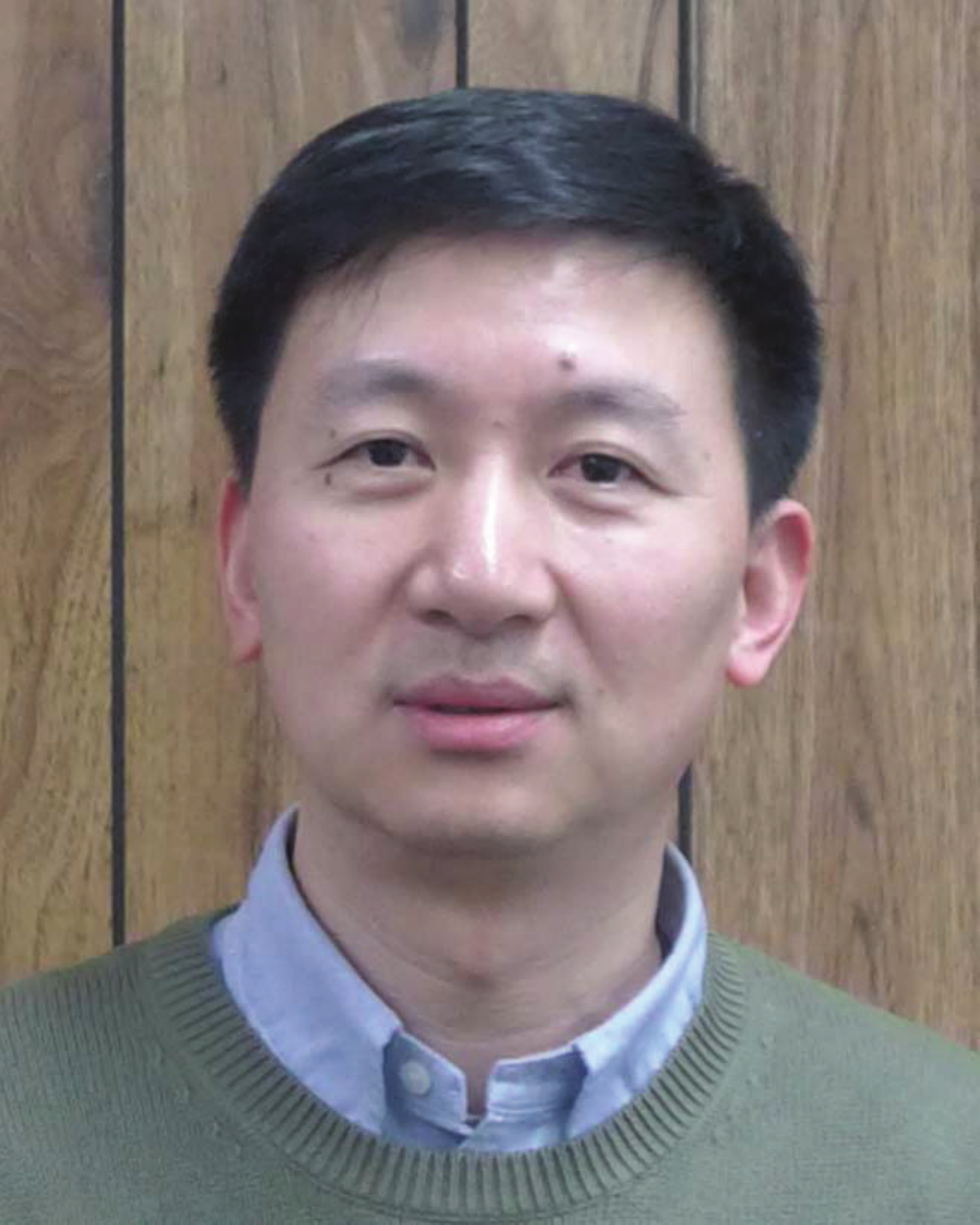}}]{Yudong Yao} (S'88-M'88-SM'94-F'11) received B.Eng. and M.Eng. degrees from Nanjing University of Posts and Telecommunications, Nanjing, in 1982 and 1985, respectively, and a Ph.D. degree from Southeast University, Nanjing, in 1988, all in electrical engineering. He was a visiting student at Carleton University, Ottawa, in 1987 and 1988. Dr. Yao has been with Stevens Institute of Technology, Hoboken, New Jersey, since 2000 and is currently a professor and department director of electrical and computer engineering. He is also a director of the Stevens' Wireless Information Systems Engineering Laboratory (WISELAB). Previously, from 1989 to 2000, Dr. Yao worked for Carleton University, Spar Aerospace Ltd., Montreal, and Qualcomm Inc., San Diego. He has been active in a nonprofit organization, WOCC, Inc., which promotes wireless and optical communications research and technical exchange. He served as WOCC president (2008-2010) and chairman of the board of trustees (2010-2012). His research interests include wireless communications and networking, cognitive radio, machine learning, and big data analytics. He holds one Chinese patent and thirteen U.S. patents. Dr. Yao was an Associate Editor of IEEE Communications Letters (2000-2008) and IEEE Transactions on Vehicular Technology (2001-2006) and an Editor for IEEE Transactions on Wireless Communications (2001-2005). He was elected an IEEE Fellow in 2011 for his contributions to wireless communications systems and was an IEEE ComSoc Distinguished Lecturer (2015-2018). In 2015, he was elected a Fellow of National Academy of Inventors.
\end{IEEEbiography}
\begin{IEEEbiography}[{\includegraphics[width=1in,height=1.25in,clip,keepaspectratio]{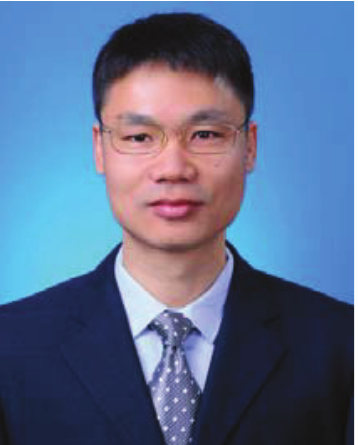}}]{Yueyang Teng} received his Bachelor and Master degrees from the Department of Applied Mathematics, Liaoning Normal University and Dalian University of Technology, China, in 2002 and 2005, respectively. From 2005 to 2013, he was a software engineer at Neusoft Positiron Medical Systems Co., Ltd. In 2013, he received his doctorate degree in Computer Software and Theory, Northeastern University. Since 2013, he has been a lecturer at the Sino-Dutch Biomedical and Information Engineering School, Northeastern University. His research interests include image processing and machine learning.
\end{IEEEbiography}
\end{document}